\documentclass{article}
\PassOptionsToPackage{table}{xcolor}
\usepackage{iclr2027_conference,times}

\iclrfinalcopy 

\usepackage[utf8]{inputenc} 
\usepackage[T1]{fontenc}    
\usepackage[table]{xcolor}  
\usepackage[colorlinks,breaklinks]{hyperref} 
\hypersetup{
    citecolor=[RGB]{50,100,170},
    linkcolor=[RGB]{50,100,170},
    urlcolor=[RGB]{255,102,178}}
\usepackage{url}            
\usepackage{booktabs}       
\usepackage{multirow}       
\usepackage{graphicx}       
\usepackage{capt-of}        
\usepackage{wrapfig}        
\usepackage{needspace}      
\usepackage{amsmath}        
\usepackage{amsfonts}       
\usepackage{algpseudocode}
\usepackage{nicefrac}       
\usepackage{microtype}      
\usepackage[nameinlink,capitalize]{cleveref}
\newcounter{algorithm}
\renewcommand{\thealgorithm}{\arabic{algorithm}}
\crefname{algorithm}{Algorithm}{Algorithms}
\Crefname{algorithm}{Algorithm}{Algorithms}
\crefname{table}{Tab.}{Tabs.}
\Crefname{table}{Tab.}{Tabs.}
\crefname{section}{Sec.}{Secs.}
\Crefname{section}{Sec.}{Secs.}
\crefname{subsection}{Sec.}{Secs.}
\Crefname{subsection}{Sec.}{Secs.}
\crefname{appendix}{Appendix}{Appendices}
\Crefname{appendix}{Appendix}{Appendices}
\definecolor{baselinegray}{HTML}{EEEEEE}
\definecolor{tsinghuapurple}{HTML}{660874}
\newcommand{\perfscore}[2]{\textcolor{tsinghuapurple!#1!white}{#2}}
\definecolor{gaincolor}{HTML}{E6F4EA}
\definecolor{dropcolor}{HTML}{FCE8E6}
\definecolor{gaintext}{HTML}{2B7A78}
\definecolor{droptext}{HTML}{A85D68}
\definecolor{corecolor}{HTML}{EAF3FF}
\definecolor{coreblue}{HTML}{1F4E79}

\newcommand{\rec}[1]{\textcolor{gaintext}{#1}}
\newcommand{\fail}[1]{\textcolor{droptext}{#1}}

\newcommand{\ours}{TRD}
\newcommand{\oursfull}{Trajectory-Refined Distillation}
\title{\oursfull{}}

\author{%
  Li Jiang$^{1,2}$\thanks{Equal contribution. Correspondence to \texttt{li.jiang3@mail.mcgill.ca}}, \ Haoran Xu$^{3}$\footnotemark[1], \ Yichuan Ding$^{1}$, \ Amy Zhang$^{3}$ \\
  $^{1}$McGill University, $^{2}$Mila Quebec AI Institute, $^{3}$UT Austin
}

\begin{document}

\maketitle
\lhead{Preprint}

\begin{abstract}
On-policy distillation (OPD) has become a central post-training tool for large language models (LLMs), providing dense per-token teacher supervision along the student’s own rollouts. In this work, we identify a common structural cause underlying OPD, which we call \emph{prefix failure}. Under prefix failure, dense per-token supervision induces a bimodal teacher mixture and fragmented gradients that token-level loss truncation or reweighting fail to address. We thus propose Trajectory-Refined Distillation (TRD), a trajectory-level correction method that revises the student’s rollout under the teacher guidance while within on-policy support. By correcting problematic prefixes before distillation, TRD mitigates prefix failure at its source. Moreover, TRD improves exploration by exposing the student to alternative valid derivations under teacher guidance, even when the original rollouts are already correct. TRD can also be applied to on-policy self-distillation (OPSD), a parameter-sharing variant that uses the student model conditioned on privileged information as the teacher. TRD improves average math accuracy while remaining competitive on code. Under OPSD, it solves $9/23$ AMOBench questions left unsolved by the Qwen3-8B base model, nearly doubling the strongest dense-KL baseline.
Code is available at this \href{https://github.com/louieworth/trd}{link}.
\end{abstract}

\begin{center}
\centering
\begin{minipage}[t]{0.43\textwidth}
\vspace{0.8em}
\centering
\includegraphics[width=\linewidth]{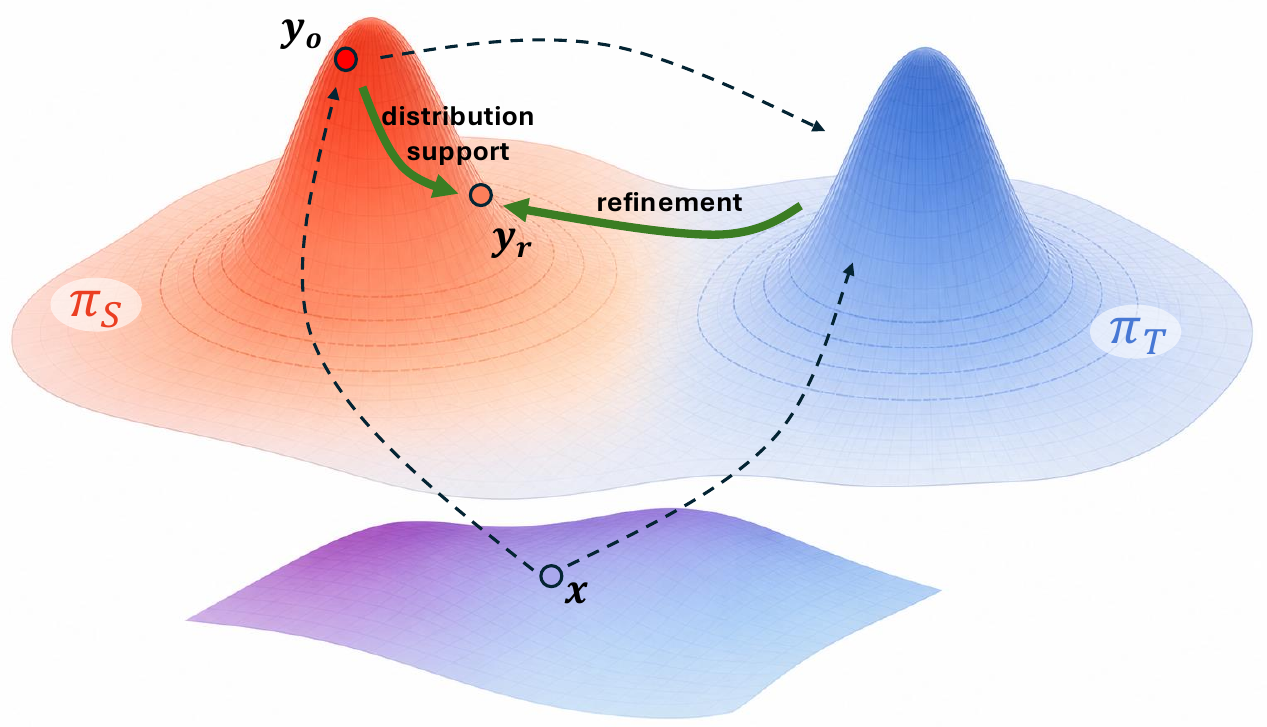}
\end{minipage}\hfill
\begin{minipage}[t]{0.5\textwidth}
\vspace{0pt}
\centering
\includegraphics[width=\linewidth]{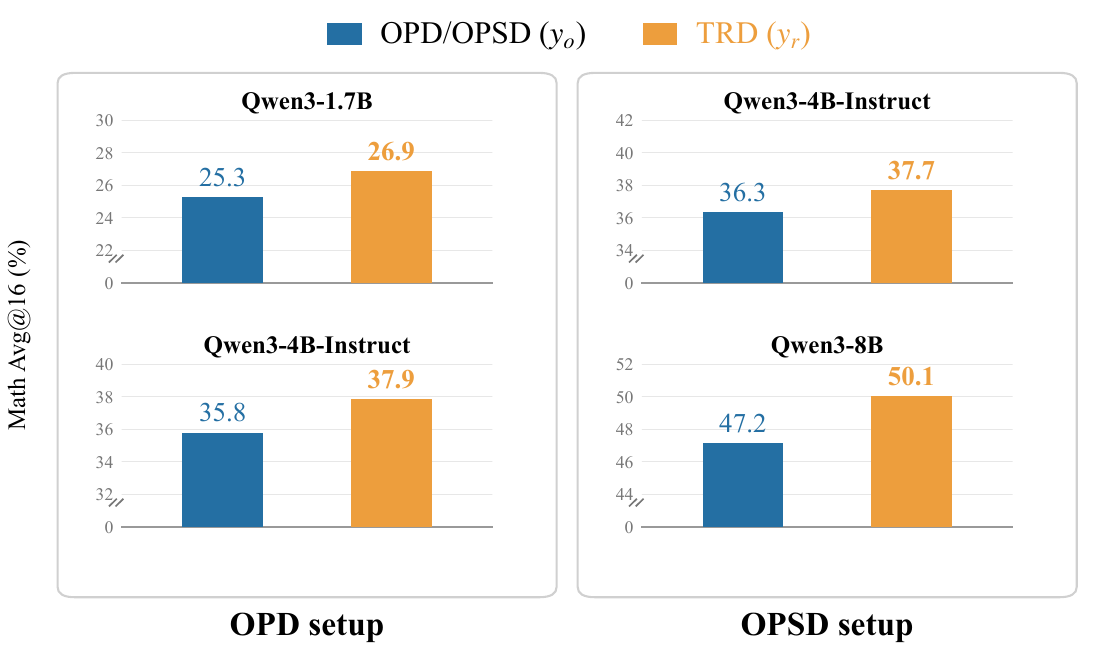}
\end{minipage}
\captionof{figure}{\textbf{Left}: \ours{} refines student-generated trajectories $y_o$ into improved trajectories $y_r$, which are then used for distillation. 
\textbf{Right}: Mean Avg@16 across the five math benchmarks for OPD/OPSD and \ours{} under different base models. }
\label{fig:intro-overview}
\end{center}
\section{Introduction}
\label{sec:intro}

On-policy distillation (OPD), which computes per-token teacher supervision along the student's own rollouts, has quickly secured a place in modern large language model (LLM) post-training \citep{gu2024minillm,agarwal2024policy,lu2025onpolicy}. Recent industry releases including Qwen3~\citep{yang2025qwen3}, DeepSeek-v4~\citep{deepseek2026v4}, MiMo-v2~\citep{xiao2026mimo}, and GLM-5~\citep{zeng2026glm} all incorporate an OPD stage alongside supervised fine-tuning (SFT) and reinforcement learning with verifiable rewards (RLVR). 
While OPD typically relies on a distinct teacher model to supervise training, On-policy self-distillation (OPSD) offers a lightweight alternative: it acts as the parameter-sharing variant of OPD in which the teacher and student are the same model under different contexts. The student is conditioned only on the problem statement, while the teacher is additionally conditioned on privileged information, such as the ground-truth answer~\citep{opsd2025,hubotter2026reinforcement,shenfeld2026self}.

Despite these successes, recent studies show that vanilla OPD/OPSD recipes often fall short of their promise, exhibiting failure modes that turn the supervision signal noisy or totally uninformative~\citep{fu2026revisiting,xu2026tip}. 
Existing remedies address these issues at the token-loss level, reweighting or clipping per-token contributions while leaving the sampled trajectory itself unchanged. For example, \citet{opsd2025} clip per-token losses above a fixed threshold to suppress destabilizing high-KL tokens.
Unfortunately, by acting only at the per-token loss level, these interventions fail to mitigate \emph{prefix failure}, an inherent limitation we formalize in this work. Prefix failure happens when the student's rollout takes a wrong reasoning path and almost no continuation of that prefix can reach the correct solution without backtracking or reflection. 
Under prefix failure, the per-token teacher distribution becomes a bimodal mixture between continuing the failed prefix and pivoting toward the correction (\cref{subsec:mixture_of_distribution}). Even with an ideal teacher, evaluating per-token KL along the student's \emph{frozen rollout} fragments the gradient and yields supervision pairs that diverge from the correction path itself (\cref{subsec:perfect_teacher}). 
Recovery therefore requires a \emph{trajectory-level} improvement; prior per-token approaches operate at the wrong scale, i.e., token-level, leaving the failed prefix structurally intact.

To mitigate the prefix failure, we propose \oursfull{} (\ours{}), a simple yet effective trajectory-level refinement strategy that retains on-policy support under teacher guidance.
Concretely, for each problem, we first sample a raw on-policy rollout and then prompt the teacher model to produce a refined version of that rollout. The refined trajectory is then used as supervision for subsequent distillation (\cref{sec:method}). TRD naturally extends to self-distillation, where the teacher additionally receives the reference solution.
We empirically validate \ours{} in both OPD and OPSD on five competition-math benchmarks using Qwen3 models at multiple scales; we additionally evaluate code generation on three code generation tasks for OPD and OPSD variants. On math, \ours{} achieves the best average performance against baselines across settings, and on code it remains competitive while revealing benchmark-specific trade-offs. Under OPSD, Qwen3-8B with \ours{} also solves $9$ of the $23$ AMOBench questions that the base model leaves unsolved in $16$ attempts, nearly doubling the strongest dense-KL baseline's Pass@16 on this subset (\cref{app:test-rollout}).

\section{Related Work}
\label{sec:related}

\paragraph{On-Policy Distillation.} 
On-policy distillation (OPD) for post-training LLMs traces back to classical knowledge distillation~\citep{hinton2015distilling} and replaces the fixed-corpus targets there with per-token teacher supervision computed along the student's own rollouts. OPD couples with the on-policy sampling property and dense token-level learning signal by token-level KL loss~\citep{gu2024minillm,agarwal2024policy,song2026survey,yang2025qwen3,deepseek2026v4,xiao2026mimo,zeng2026glm}. By contrast, RLVR offers only a sparse trajectory-level reward that scales poorly when most rollouts fail the verifier, whereas SFT relies on off-policy reference data and forfeits the on-policy structure that drives compute-efficient learning \citep{lu2025onpolicy}. 

\paragraph{On-policy Self-distillation.}
On-policy self-distillation (OPSD) is a special case of OPD that instantiates teacher and student from the same model under different privileged contexts, thereby enabling self-improvement without a separate teacher model~\citep{opsd2025,hubotter2026reinforcement,shenfeld2026self}. The privileged context typically can be involved in reference solutions, feedback, knowledge and experiences, or other auxiliary information that is unavailable to the student \citep{shi2026experiential,ye2026policy,penaloza2026privileged,wang2026skill,stein2026gates}. Operating within the OPD paradigm, OPSD inherits most of the same failure modes, e.g., the privileged supervision can collapse into vacuous guidance as training progresses. A similar concern appears in \citet{opsd2025}, who clip unusually large per-token losses to suppress unreliable learning signals and stabilize training. 

\paragraph{Common Failure Mode and Fix.}
Recent studies report that those vanilla distillation methods often underperform in practice and exhibit a range of failure modes, including mode collapse, trajectory inflation, and supervision signals that vanish or even actively mislead the student and more \citep{fu2026revisiting,xu2026tip,luo2026demystifying,yang2026selfdistilledrlvr,kim2026does,li2026unifying,xu2026paced,song2026survey}. Most of these failure modes are addressed through token-level dense-KL loss interventions. For example, \citet{fu2026revisiting} find that the teacher distribution under OPD can be dominated by a small number of high-loss tokens, and propose top-$K$ truncation to restrict supervision to high-confidence tokens. \citet{xu2026tip} upweights informative tokens, e.g., tokens with low student entropy but high teacher–student divergence, to achieve better results.


\paragraph{Concurrent work on prefix failure.}
Relay-OPD uses brief teacher takeovers when the teacher favors correction and the student continues along a failed prefix~\citep{xu2026relay}. Motab backtracks to a safe prefix before generating a corrective suffix~\citep{wang2026motab}. OLIVE trains on teacher continuations of freshly sampled student prefixes using cross-entropy~\citep{wang2026olive}. For multi-turn agents, ReOPD replays teacher trajectories as prefixes and favors early interaction steps to limit accumulated prefix errors~\citep{liao2026reopd}. Our analysis explains how prefix failure produces bimodal teacher targets and fragmented gradients; TRD refines complete student rollouts before KL distillation in both OPD and OPSD.

\section{Preliminaries}
\label{sec:pre}
\paragraph{On-policy Distillation.}
Knowledge distillation trains a student model $\pi_S$ to match the output distribution of a teacher $\pi_T$, beyond the hard reference label~\citep{hinton2015distilling}. In on-policy distillation (OPD) of LLMs, supervision is computed on trajectories sampled from the current student rather than on fixed expert prefixes~\citep{gu2024minillm,agarwal2024policy,lu2025onpolicy}. Given a prompt $x\sim\mathcal{D}$ from the training dataset, the student samples an autoregressive rollout $y\sim \pi_\theta(\cdot\mid x)$. The teacher $\pi_T(\cdot\mid x,y_{<t})$ is then evaluated on the student-visited prefixes $y_{<t}$, producing a dense token-level learning signal. A representative OPD objective minimizes the per-token reverse KL divergence between student and teacher,
\begin{equation}
        \mathcal{L}(\theta)
    =
    \mathbb{E}_{x\sim\mathcal{D},\,y\sim \pi_\theta(\cdot\mid x)}
    \left[\frac{1}{|y|}\sum_{t=1}^{|y|}
    D\big( \pi_\theta(\cdot\mid x,y_{<t})\,\big\|\, \pi_T(\cdot\mid x,y_{<t})\big)\right].
\label{eq:opd_loss}
\end{equation}


\paragraph{On-policy Self-Distillation.}
On-policy self-distillation (OPSD) removes the need for a separate teacher by embodying teacher and student policies from the same model under different contexts~\citep{opsd2025,hubotter2026reinforcement,shenfeld2026self}. Given a problem-solution pair $(x,y^\star)\sim\mathcal{D}$, the teacher policy receives privileged information such as the reference answer or reasoning trace and is evaluated as $\pi_T=\pi_\theta(\cdot\mid x,y^\star,y_{<t})$, with teacher and student sharing the parameters $\theta$ of the same model. The standard OPSD loss is
\begin{equation}
\label{eq:opsd_loss}
    \mathcal{L}(\theta)
    =
    \mathbb{E}_{(x,y^\star)\sim \mathcal{D},\,y\sim \pi_\theta(\cdot\mid x)}
    \left[
    \frac{1}{|y|}\sum_{t=1}^{|y|}
    D\big(
    \pi_\theta(\cdot\mid x,y_{<t})
    \,\big\|\, 
    \mathrm{sg}\!\left[\pi_\theta(\cdot\mid x,y^\star,y_{<t})\right] 
    \big)
    \right],
\end{equation}
where $\mathrm{sg}[\cdot]$ denotes the stop-gradient operator applied to the teacher branch. The choice of divergence is itself a design. Reverse KL is mode-seeking and is generally preferred for generative language-model distillation since it discourages the student from assigning probability to low-probability regions of the teacher, whereas forward KL is mode-covering~\citep{gu2024minillm,lu2025onpolicy,opsd2025,jin2026entropy,jang2026stable}. In practice, the full vocabulary $\mathcal{V}$ is widely adopted to reduce variance and stabilize gradient estimates relative to single-sample Monte Carlo estimation \citep{opsd2025,yang2025qwen3}. Having full access to both distributions makes the KL direction independent of the sampling distribution; see \citet{jin2026entropy} for detailed discussion.
\section{Prefix Failure in Token-Level On-Policy Distillation}
\label{sec:prefix_failure}



\subsection{Mixture of Distribution}
\label{subsec:mixture_of_distribution}
\begin{figure}[!htbp]
\centering
\includegraphics[width=\linewidth]{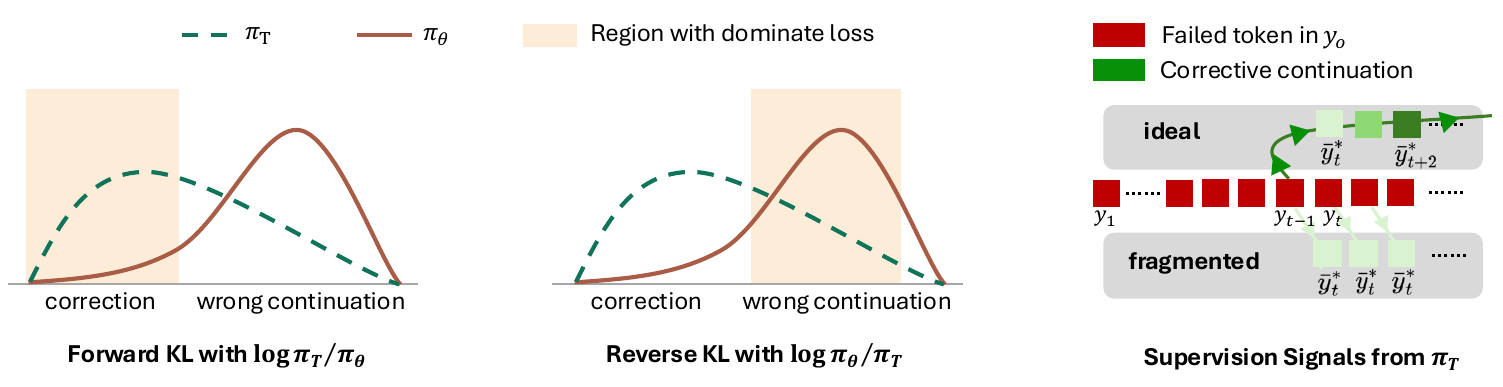}
\caption{\textbf{Left and Middle:} Under prefix failure, the teacher distribution becomes a mixture with two modes. By their respective mode-covering and mode-seeking properties, forward KL is dominated by the correction-onset region, while reverse KL is dominated by the wrong-continuation region. \textbf{Right:} Supervision signal provided by the teacher under prefix failure. }
\label{fig:prefix-failure}
\end{figure}

Dense per-token KL relies on the teacher providing a reliable supervisory signal at every position along the student rollout. When the teacher faces a wrong reasoning path rolled out by the students, the supervision can become unreliable. Let $\mathsf{F}(y_{o,<t})$ denote the \emph{prefix-failure} event: the prefix $y_{o,<t}$ contains reasoning errors that contradict $y^\star$ and cannot be extended to $y^\star$ without retraction or contradiction. Whenever $\mathsf{F}(y_{o,<t})$ holds, the teacher becomes a mixture distribution with two modes: one that continues the failed prefix $y_{o,<t}$ for sequence consistency and another that pivots back toward $y^\star$. This mixture structure turns the supposedly dense supervision into a noisy or even adversarial signal. A complementary failure mode arises on degenerate prefixes (e.g., repetition loops), where the teacher instead remains locally aligned with the student and the guidance signal vanishes entirely~\citep[Figure 3]{fu2026revisiting}. Here, we focus on the supervision mismatch caused by prefix failure in dense on-policy distillation. SFT uses fixed reference trajectories. RLVR updates sampled trajectories using verifier-based rewards rather than dense teacher distributions at each prefix.

The choice of KL direction interacts with prefix failure asymmetrically. Recovering from $\mathsf{F}(y_{o,<t})$ requires a corrective continuation $\rec{\bar{y}^{\star}_{\geq t} = (\bar{y}^{\star}_t, \bar{y}^{\star}_{t+1}, \ldots, \bar{y}^{\star}_{t+k-1})}$\footnote{$\bar{y}^{\star}_{\geq t}$ is \emph{not} the literal suffix of $y^\star$, but any correct continuation conditioned on $y_{<t}$, i.e., $\operatorname{Verify}(y_{<t} \cdot \bar{y}^{\star}_{\geq t},\, y^\star) = 1$.}, generated autoregressively along the correction path, where $\bar{y}^\star_t$ is a correction-onset token ($\bar{y}^\star_t \in \{\texttt{Wait}, \texttt{Actually}, \ldots\}$) and $\bar{y}^\star_{t+1}, \ldots, \bar{y}^\star_{t+k-1}$ continue the recovery, with $k$ denoting the length of this continuation.
Under $\mathsf{F}(y_{o,<t})$, the ideal teacher partially shifts mass from the natural continuation toward $\bar{y}^\star_t$, while the student with high probability remains anchored on the wrong continuation.
Forward KL $D(\pi_T \| \pi_\theta)$, weighted by $\pi_T$, is dominated by the correction-onset region; its mode-covering nature forces the student onto this out-of-distribution (OOD) mode. This can destabilize training and, in the worst case, lead to mode collapse \citep{opsd2025}.
In contrast, reverse KL $D(\pi_\theta \| \pi_T)$ is weighted by $\pi_\theta$, so its mode-seeking behavior makes the loss dominated by the wrong-continuation region where the student already places high mass. Since $\pi_\theta(\bar{y}^\star_t)$ is small, the correction signal has limited effect, so updates concentrate on the failing trajectory rather than the recovery tokens.

Prior work has identified several failure modes of dense token-level KL training and proposed loss-level remedies. These failures are consistent with the prefix-failure mechanism above, even when not explicitly framed this way. Under forward KL, teacher-weighted correction or OOD modes can already dominate the loss, so \citet{opsd2025} clip per-token losses to cap unstable high-KL terms. Under reverse KL, the correction mode is instead underweighted by the student-weighted objective; accordingly, \citet{fu2026revisiting} use teacher top-$K$ truncation, and \citet{xu2026tip} reweight losses by entropy and student-teacher disagreement to recover informative teacher-preferred tokens. Thus, these remedies control token-level dominance in opposite directions: clipping suppresses overly dominant forward-KL terms, while truncation or reweighting amplifies underweighted reverse-KL teacher modes. However, all of them leave the failed prefix unchanged.

\subsection{Can a Perfect Teacher Recover the Correction Path?}
\label{subsec:perfect_teacher}
Even granting an ideal teacher, dense per-token KL is structurally limited because it is a \emph{post-hoc} per-token objective evaluated along the student's rollout. To make this precise, we trace the per-token OPD loss back to its sequence-level origin, which makes the underlying mechanism more transparent. Define the per-token log-ratio $\delta_t \!:=\! \log \pi_\theta(y_t\mid x,y_{<t}) - \log \pi_T(y_t\mid x,y_{<t})$. Differentiating the sequence-level reverse KL without length normalization (full derivation in \cref{app:opd-derivation}) yields the policy-gradient form
\begin{equation*}
    \nabla_{\theta} \mathcal{J}_{\mathrm{seq}}(\theta)
    \;=\;
    \mathbb{E}_{\substack{x\sim\mathcal{D}\\ y\sim \pi_\theta(\cdot\mid x)}}
    \!\left[
    \sum_{t=1}^{|y|}
    \bigg(
    \delta_t
    \;+\; \sum_{t'=t+1}^{|y|} \delta_{t'}
    \bigg)\,
    \nabla_{\theta} \log \pi_\theta(y_t\mid x,y_{<t})
    \right],
\end{equation*}
where $-\delta_t$ acts as a token-level return. In practice, however, standard OPD implementations~\citep{lu2025onpolicy,yang2025qwen3} do not optimize this sequence-level objective directly; instead, they retain only the immediate log-ratio at each position, yielding the per-token surrogate
\begin{equation}
\label{eq:opsd_grad}
    g_{\mathrm{tok}}(\theta)
    \;:=\;
    \mathbb{E}_{\substack{x\sim\mathcal{D}\\ y\sim \pi_\theta(\cdot\mid x)}}
    \!\left[
    \sum_{t=1}^{|y|}
    \delta_t
    \nabla_{\theta} \log \pi_\theta(y_t\mid x,y_{<t})
    \right].
\end{equation}
For comparison, an ideal teacher autoregressively unfolds $\rec{\bar{y}^\star_{\geq t}}$ along the correction path, delivering the supervision pairs
\begin{equation*}
\big\{(y_{o,<t},\, \rec{\bar{y}^\star_t}),\;\; \big((y_{o,<t}, \rec{\bar{y}^\star_t}),\, \rec{\bar{y}^\star_{t+1}}\big),\;\; \ldots,\;\; \big((y_{o,<t}, \rec{\bar{y}^\star_{t:t+k-2}}),\, \rec{\bar{y}^\star_{t+k-1}}\big)\big\},
\end{equation*}
in which the context grows along the correction path itself. Define an
illustrative correction-token update signal
\begin{equation*}
u_{\text{ideal}} \;:=\; -\sum_{i=1}^{k} \delta_i^{\text{ideal}} \cdot \nabla_\theta \log \pi_\theta\!\big(\rec{\bar{y}^\star_{t+i-1}} \,\big|\, x,\; y_{o,<t},\, \rec{\bar{y}^\star_{t:t+i-2}}\big),
\end{equation*}
where $\bar{y}^\star_{t:t-1}$ is empty and $\delta_i^{\text{ideal}} := \log \pi_\theta(\rec{\bar{y}^\star_{t+i-1}} \mid x, y_{o,<t}, \rec{\bar{y}^\star_{t:t+i-2}}) - \log \pi_T(\rec{\bar{y}^\star_{t+i-1}} \mid x, y_{o,<t}, \rec{\bar{y}^\star_{t:t+i-2}})$. Yet under dense KL the contexts are dictated by the \emph{frozen} student trajectory, not by the unfolding correction. At position $t$, the teacher conveys $\rec{\bar{y}^\star_t}$ given $y_{o,<t}$, and the student updates its parameters to favor $\rec{\bar{y}^\star_t}$. At position $t+1$, however, the teacher's supervision is conditioned on $\fail{y_{o,<t+1}} = (y_{o,<t}, \fail{y_{o,t}})$ (the original failed trajectory's own continuation), not on the correction path $(y_{o,<t}, \rec{\bar{y}^\star_t})$. If the teacher repeatedly favors the same correction-onset token on these failed prefixes, the selected supervision pairs form the fragmented sequence
\begin{equation*}
\big\{(y_{o,<t},\, \rec{\bar{y}^\star_t}),\;\; (\fail{y_{o,<t+1}},\, \rec{\bar{y}^\star_t}),\;\; \ldots,\;\; (\fail{y_{o,<t+k-1}},\, \rec{\bar{y}^\star_t})\big\},
\end{equation*}
in which the context grows along the wrong continuation while the target
remains stuck at $\bar{y}^\star_t$. The analogous update signal
\begin{equation*}
u_{\text{frag}} \;:=\; -\sum_{i=0}^{k-1} \delta_i^{\text{frag}} \cdot \nabla_\theta \log \pi_\theta\!\big(\rec{\bar{y}^\star_t} \,\big|\, x,\; \fail{y_{o,<t+i}}\big),
\end{equation*}
uses $\delta_i^{\text{frag}} := \log \pi_\theta(\rec{\bar{y}^\star_t} \mid x, \fail{y_{o,<t+i}}) - \log \pi_T(\rec{\bar{y}^\star_t} \mid x, \fail{y_{o,<t+i}})$. These illustrative signals compare selected (context, token) pairs;
neither is a full KL gradient, and the negative sign denotes the update direction.
When $\bar{y}^\star_t \ne y_{o,t}$, the two pair sets share only their first element $(y_{o,<t}, \rec{\bar{y}^\star_t})$;
beyond it, $u_{\text{ideal}}$ follows
$(y_{o,<t}, \rec{\bar{y}^\star_t}, \rec{\bar{y}^\star_{t+1}}, \ldots)$ while $u_{\text{frag}}$ follows
$(y_{o,<t}, \fail{y_{o,t}}, \fail{y_{o,t+1}}, \ldots)$. The two
trajectories diverge after a single step and never re-intersect:
\begin{equation*}
\underbrace{\big\{((y_{o,<t}, \rec{\bar{y}^\star_{t:t+i-2}}),\, \rec{\bar{y}^\star_{t+i-1}})\big\}_{i=1}^{k}}_{\text{required by recovery}} \;\;\cap\;\; \underbrace{\big\{(\fail{y_{o,<t+i}},\, \rec{\bar{y}^\star_t})\big\}_{i=0}^{k-1}}_{\text{delivered}} \;=\; \big\{(y_{o,<t},\, \rec{\bar{y}^\star_t})\big\}.
\end{equation*}
In this example, the privileged information $y^\star$ stays trapped in per-position marginals. Dense KL keeps recommending $\rec{\bar{y}^\star_t}$ on ever-deepening wrong-continuation contexts but cannot supervise the multi-step unfolding of $\rec{\bar{y}^\star_{\geq t}}$, and loss-level interventions only reweight terms within the visited pair set $\{(\fail{y_{o,<t+i}}, \rec{\bar{y}^\star_t})\}$ rather than move supervision onto the correction-path pair set. The $u_{\text{frag}}$ signal is not useless, however; biasing the student toward $\rec{\bar{y}^\star_t}$ can trigger self-correction, though unfolding the full $\rec{\bar{y}^\star_{\geq t}}$ is bounded by the student's capacity. Our method \ours{} (\cref{sec:method}) instead evaluates per-token KL along a teacher-refined trajectory, allowing supervision contexts to follow the refined path rather than the frozen failed prefix. 

Complementary training diagnostics are reported in \cref{subsec:exp_prefix}. They characterize teacher-student discrepancies on raw and refined trajectories; the structural argument above concerns which supervision contexts are visited, rather than the absolute magnitude of the KL loss.

\section{\oursfull{}}
\label{sec:method}

The previous section identifies \emph{prefix failure} as one of the central bottlenecks in OPD. With prefix failure, dense supervision becomes noisy and can even fail to guide the student. Most existing mitigations operate through loss design and leave the offending prefix unchanged. Directly optimizing prefix failure is intractable: whether a prefix has failed is only revealed after the full trajectory is verified, while locating the failure index would require searching $\mathcal{O}(|\mathcal{V}|^k)$ continuations of length $k$. We therefore relax the target to a trajectory-level surrogate that maximizes the expected verifier-pass rate over the dataset $\mathcal{D}$ under the support constraint of $\pi_\theta$:
\begin{equation}
\label{eq:prefix-refinement-objective}
\max _q \mathbb{E}_{\left(x, y^{\star}\right) \sim \mathcal{D}}\left[\operatorname{Pr}_{y \sim q(\cdot \mid x, \cdot)}\left\{\operatorname{Verify}\left(y, y^{\star}\right)=1\right\}\right] \quad \text { s.t. }\quad  \pi_\theta(y \mid x)>0
\end{equation}

We note that this is a \emph{trajectory-level} distribution support constraint and the optimization objective form mirrors a standard RLVR objective. We emphasize, however, that this objective is not optimized directly in the OPD update; rather, it defines an upstream trajectory-construction task before the standard OPD optimization over $\theta$, namely to construct trajectories that attain higher verifier-pass rates while remaining within the current student's support. These trajectories then serve as the supervision for the subsequent OPD update through the standard distribution-matching loss in \cref{eq:opd_loss}. 

Two extreme choices of $q$ illustrate the tension between objective and constraint. \textbf{\textsc{(i)}} Setting $q(\cdot \mid  x, \cdot)=\pi_\theta(\cdot\mid x)$ fails back to standard OPD. The on-policy constraint is satisfied by construction, but this choice fails to mitigate \emph{prefix failure} beyond current OPD algorithms, since the supervision trajectories are still drawn from $\pi_\theta$. Reject sampling refines this choice by drawing rollouts and retaining only the verifier-passing ones \citep{brown2024large,stein2026gates}, straining inference budgets linearly and yielding nothing on questions the student cannot solve. \textbf{\textsc{(ii)}} Setting $q(\cdot \mid  x, \cdot)=\pi^\star$, i.e., the expert policy that produces $y^\star$, attains $\operatorname{Verify}=1$ exactly but generally violates the on-policy support constraint, so this choice falls outside the feasible set and breaks the on-policy character that OPD depends on.

To move beyond the two extremes above, we propose \emph{\oursfull{}} (\ours{}), which operates at the trajectory level to optimize \cref{eq:prefix-refinement-objective} by first drawing a raw on-policy rollout $y_o \sim \pi_\theta(\cdot\mid x)$, then asking the teacher to construct a refined trajectory $y_r$ via:
\begin{equation}
y_r \sim q(\cdot\mid x,\cdot)\;:=\;\pi_T(\cdot\mid x,y_o).
\label{eq:rsd}
\end{equation}
In OPSD, the same backbone implements this teacher query by additionally conditioning on the reference solution $y^*$: $y_r \sim q(\cdot\mid x,\cdot):=\mathrm{sg}[\pi_\theta(\cdot\mid x,y_o, y^*)]$.
Crucially, conditioning on $y_o$ anchors $y_r$ to the reasoning patterns $\pi_\theta$ has already demonstrated, i.e., within the policy support, while $\pi_T$ rewrites the erroneous portions to directly mitigate prefix failure. Findings in SFT also support the value of student-familiar responses~\citep{ren-etal-2024-learn,yang-etal-2024-self}. To our knowledge, this is the first trajectory-level optimization design that explicitly targets \cref{eq:prefix-refinement-objective} while respecting the on-policy constraint without prohibitive additional compute overhead. A refined trajectory $y_r$ is then used as supervision for the subsequent OPD update. Per-token KL along $y_r$ reduces exposure to the bimodal teacher mixture (\cref{subsec:mixture_of_distribution}) and evaluates supervision on refined prefixes (\cref{subsec:perfect_teacher}), rather than the raw rollout $y_o$ at higher risk of prefix failure.

\begin{wrapfigure}{r}{0.38\textwidth}
\vspace{-1em}
\centering
\begingroup
\setlength{\tabcolsep}{0pt}
\begin{minipage}{0.38\textwidth}
\hrule height 0.8pt
\vspace{0.35em}
\refstepcounter{algorithm}\label{alg:prefix-refinement}\textbf{Algorithm \thealgorithm: \oursfull{}}
\vspace{0.25em}
\hrule
\vspace{0.35em}
\begin{algorithmic}[1]
\Require minibatch $\mathcal{B}\subset\mathcal{D}$, $\pi_\theta$, $\pi_T$
\For{$x\in\mathcal{B}$}
    \State $y_o\sim\pi_\theta(\cdot\mid x)$
    \State $y_r\sim\pi_T(\cdot\mid x,y_o)$ by \cref{eq:rsd}
    \State Update $ \theta $ by \cref{eq:prefix_refinement_loss} on $y_r$
\EndFor
\end{algorithmic}
\vspace{0.25em}
\hrule height 1pt
\end{minipage}
\endgroup
\end{wrapfigure}

\ours{} also boosts the student's exploration beyond standard OPD. On a correct $y_o$, standard OPD can still provide useful supervision, but the update is evaluated only on the prefixes of that sampled solution. Refinement additionally samples $y_r$ from $\pi_T(\cdot\mid x,y_o)$ and can expose alternative valid derivations of the same answer, i.e., paths suggested by the teacher but rarely sampled from $\pi_\theta(\cdot \mid x)$ alone, thereby expanding the set of correct reasoning trajectories the student is supervised on. \ours{} therefore adds value in both regimes: mitigating failed prefixes when $y_o$ exhibits prefix failure, and broadening the supervision distribution when $y_o$ already succeeds. The training-data analysis in \cref{subsec:trajectory-analysis} supports this interpretation (e.g., the correct subset of $y_r$ exhibits a low-length mode absent in $y_o$) and is complemented by the test-time coverage analysis in \cref{subsec:exploration}.

Concretely, given $y_r$, we instantiate the distillation loss as the forward KL with full vocabulary matching over $\mathcal{V}$, which provides mode-covering supervision and stabilizes gradient estimates:
\begin{equation}
\begin{aligned}
    \mathcal{L}(\theta)
    =
    \mathbb{E}_{\substack{
        x\sim\mathcal{D}\\
        y_o\sim\pi_\theta(\cdot\mid x)\\
        y_r\sim\pi_T(\cdot\mid x,y_o)
    }}
    \left[
    \frac{1}{|y_r|}
    \sum_{t=1}^{|y_r|}
    D\big(
    \mathrm{sg}\!\left[
    \pi_T(\cdot\mid x,y_o,y_{r,<t})
    \right]
    \,\big\|\,
    \pi_\theta(\cdot\mid x,y_{r,<t})
    \big)
    \right],
\end{aligned}
\label{eq:prefix_refinement_loss}
\end{equation}
The exact prompt template is given in \cref{app:refinement-prompt}; \cref{fig:intro-overview} (left) and \cref{alg:prefix-refinement} illustrate the full procedure. \Cref{subsec:exp_prefix} reports complementary training diagnostics on $y_o$ and $y_r$. These describe changes in the supervision distributions, while downstream effectiveness is evaluated in \cref{sec:exp}.
\section{Experiments}
\label{sec:exp}
We evaluate \ours{} against SFT, GRPO, three on-policy distillation variants, and SKD under two teacher-information regimes across math and code benchmarks: a separate Qwen3-8B teacher and a privileged self-teacher. We first compare Avg@16 (\cref{exp:main}) and then analyze solution discovery under increasing test-time sampling budgets (\cref{subsec:exploration}). Best-of-$N$ comparisons under OPSD then test the trajectory-construction objective (\cref{subsec:objective-validation}). Additional objective-validation results, trajectory analyses, and a comparison of the two teacher regimes are provided in \cref{app:objective-validation,subsec:trajectory-analysis,subsec:opd-opsd-trd,app:test-rollout}, with full experimental details in \cref{app:experiment-details}.

\subsection{Experiments Setup and Baselines}
\label{exp:setup}
\paragraph{Models.}
We use the Qwen3 model family~\citep{yang2025qwen3} and evaluate two teacher regimes. In the separate-teacher regime, a frozen Qwen3-8B teacher supervises Qwen3-1.7B and Qwen3-4B-Instruct-2507 students without access to the reference solution. In the privileged self-teacher regime, teacher and student share the same Qwen3-4B-Instruct-2507 or Qwen3-8B backbone, and the stop-gradient teacher branch receives the reference solution $y^\star$ as privileged context.

\paragraph{Training Datasets.}
For math, we train on the DeepScaleR math corpus~\citep{deepscaler2025} of roughly 40K problems with solutions; for code, we train on TACO~\citep{li2023taco}, an algorithmic code-generation corpus with roughly 25K training problems, with reference solutions and test cases. In the privileged self-teacher regime, privileged conditioning uses the dataset reference solution $y^\star$.

\paragraph{Baselines.}
We compare \ours{} with SFT trained on $y^\star$ and budget-matched GRPO~\citep{shao2024deepseekmath}, which optimizes verifier rewards on $y_o$. We also evaluate on-policy distillation using full-vocabulary forward KL~\citep{lu2025onpolicy}, top-$K$ reverse KL~\citep{fu2026revisiting}, and clipped forward KL~\citep{opsd2025}. We denote these variants by +OPD with a separate teacher and +OPSD with a privileged self-teacher, adding ``w/ Top-$K$'' or ``w/ Clip'' where applicable. All three variants train on raw student trajectories $y_o$. Finally, we include Speculative Knowledge Distillation (SKD)~\citep{xu2025speculative}, which uses interleaved student-teacher trajectories, denoted by $y_o + y_r$ in the tables. \ours{} trains only on refined trajectories $y_r$.

\paragraph{Evaluation.}
We evaluate on five math benchmarks, AIME24~\citep{maa2024aime}, AIME25~\citep{maa2025aime}, HMMT25~\citep{hmmt2025}, BeyondAIME~\citep{seed2025beyondaime}, and AMOBench~\citep{amobench2025}. For code generation, we use HumanEval+ and MBPP+~\citep{evalplus}, and LiveCodeBench v6~\citep{jain2024livecodebench}. Our primary metric is Avg@16. For each test question, we sample $K{=}16$ completions and grade each with an external verifier. Avg@16 averages their binary correctness scores within each question and then across the test set, estimating single-completion accuracy. We set maximum response lengths of $38,912$ and $16,384$ tokens for math and code tasks, respectively.

\subsection{Main Experiments}
\label{exp:main}


\begin{table}[!h]
  \caption{Separate-teacher Avg@16 results (\%). The +OPD family uses a frozen Qwen3-8B teacher without access to $y^\star$. Bold marks block best. Base scores are gray; darker purple indicates higher scores among trained methods within each model block and column.}
  \label{tab:opd-avg16}
  \centering
  \scriptsize
  \setlength{\tabcolsep}{2.2pt}
  \resizebox{\textwidth}{!}{%
  \begin{tabular}{llcccccccc}
    \toprule
    Method & Traj. & \multicolumn{5}{c}{Math} & \multicolumn{3}{c}{Code} \\
    \cmidrule(lr){3-7}\cmidrule(lr){8-10}
    & & AIME24 & AIME25 & HMMT25 & BeyondAIME & AMOBench & HumanEval+ & MBPP+ & LiveCodeBench\\
    \midrule
    \multicolumn{10}{l}{\emph{Qwen3-1.7B (w/ thinking)}} \\
    \rowcolor{gray!15}
    Base & -- & \textcolor[HTML]{707070}{44.8} & \textcolor[HTML]{707070}{33.8} & \textcolor[HTML]{707070}{21.2} & \textcolor[HTML]{707070}{20.1} & \textcolor[HTML]{707070}{1.9} & \textcolor[HTML]{707070}{60.3} & \textcolor[HTML]{707070}{48.0} & \textcolor[HTML]{707070}{30.3} \\
    +SFT & $y^\star$ & \perfscore{68}{44.5} & \perfscore{58}{33.3} & \perfscore{79}{22.7} & \perfscore{84}{19.8} & \perfscore{66}{2.1} & \perfscore{40}{60.1} & \perfscore{40}{49.3} & \perfscore{40}{29.3} \\
    \rowcolor{gray!15}
    +GRPO & $y_o$ & \perfscore{87}{47.4} & \perfscore{77}{35.2} & \perfscore{90}{23.6} & \textbf{\perfscore{100}{21.4}} & \perfscore{96}{2.9} & \perfscore{81}{62.2} & \perfscore{65}{50.1} & \perfscore{77}{31.5} \\
    \arrayrulecolor{black!25}\midrule\arrayrulecolor{black}
    +OPD & $y_o$ & \perfscore{67}{44.4} & \perfscore{76}{35.1} & \perfscore{86}{23.3} & \perfscore{98}{21.2} & \perfscore{78}{2.4} & \perfscore{65}{61.4} & \perfscore{91}{50.9} & \perfscore{98}{32.8} \\
    \rowcolor{gray!15}
    +OPD (w/ Top-$K$) & $y_o$ & \perfscore{84}{46.9} & \perfscore{83}{35.8} & \perfscore{93}{23.8} & \perfscore{99}{21.3} & \textbf{\perfscore{100}{3.0}} & \perfscore{94}{62.9} & \perfscore{72}{50.3} & \perfscore{95}{32.6} \\
    +OPD (w/ Clip) & $y_o$ & \perfscore{92}{48.1} & \perfscore{69}{34.4} & \perfscore{81}{22.9} & \perfscore{94}{20.8} & \perfscore{66}{2.1} & \perfscore{88}{62.6} & \perfscore{72}{50.3} & \perfscore{92}{32.4} \\
    \rowcolor{gray!15}
    +SKD & $y_o + y_r$ & \perfscore{40}{40.2} & \perfscore{40}{31.5} & \perfscore{40}{19.6} & \perfscore{40}{15.5} & \perfscore{40}{1.4} & \perfscore{98}{63.1} & \perfscore{56}{49.8} & \perfscore{83}{31.9} \\
    \midrule
    +\ours{} (ours) & $y_r$ & \textbf{\perfscore{100}{49.4}} & \textbf{\perfscore{100}{37.5}} & \textbf{\perfscore{100}{24.4}} & \perfscore{87}{20.1} & \textbf{\perfscore{100}{3.0}} & \textbf{\perfscore{100}{63.2}} & \textbf{\perfscore{100}{51.2}} & \textbf{\perfscore{100}{32.9}} \\
    \midrule
    \multicolumn{10}{l}{\emph{Qwen3-4B-Instruct-2507}} \\
    \rowcolor{gray!15}
    Base & -- & \textcolor[HTML]{707070}{59.1} & \textcolor[HTML]{707070}{42.9} & \textcolor[HTML]{707070}{28.3} & \textcolor[HTML]{707070}{28.2} & \textcolor[HTML]{707070}{8.1} & \textcolor[HTML]{707070}{78.0} & \textcolor[HTML]{707070}{60.1} & \textcolor[HTML]{707070}{28.1} \\
    +SFT & $y^\star$ & \perfscore{48}{60.7} & \perfscore{40}{44.3} & \perfscore{40}{28.6} & \perfscore{40}{29.4} & \perfscore{40}{8.3} & \perfscore{40}{79.3} & \perfscore{40}{61.4} & \perfscore{40}{29.1} \\
    \rowcolor{gray!15}
    +GRPO & $y_o$ & \perfscore{66}{62.3} & \textbf{\perfscore{100}{48.8}} & \perfscore{78}{31.5} & \perfscore{87}{31.9} & \textbf{\perfscore{100}{11.1}} & \perfscore{61}{80.3} & \perfscore{75}{63.6} & \textbf{\perfscore{100}{32.3}} \\
    \arrayrulecolor{black!25}\midrule\arrayrulecolor{black}
    +OPD & $y_o$ & \perfscore{40}{60.0} & \perfscore{72}{46.7} & \perfscore{63}{30.4} & \perfscore{89}{32.0} & \perfscore{74}{9.9} & \perfscore{91}{81.7} & \perfscore{84}{64.2} & \perfscore{85}{31.5} \\
    \rowcolor{gray!15}
    +OPD (w/ Top-$K$) & $y_o$ & \perfscore{59}{61.7} & \perfscore{60}{45.8} & \perfscore{66}{30.6} & \perfscore{74}{31.2} & \perfscore{72}{9.8} & \perfscore{94}{81.8} & \perfscore{86}{64.3} & \perfscore{94}{32.0} \\
    +OPD (w/ Clip) & $y_o$ & \perfscore{68}{62.5} & \perfscore{52}{45.2} & \perfscore{56}{29.8} & \perfscore{96}{32.4} & \perfscore{79}{10.1} & \perfscore{94}{81.8} & \perfscore{91}{64.6} & \perfscore{94}{32.0} \\
    \rowcolor{gray!15}
    +SKD & $y_o + y_r$ & \perfscore{70}{62.7} & \perfscore{96}{48.5} & \perfscore{43}{28.8} & \textbf{\perfscore{100}{32.6}} & \perfscore{79}{10.1} & \perfscore{94}{81.8} & \perfscore{94}{64.8} & \perfscore{94}{32.0} \\
    \midrule
    +\ours{} (ours) & $y_r$ & \textbf{\perfscore{100}{65.4}} & \perfscore{88}{47.9} & \textbf{\perfscore{100}{33.2}} & \textbf{\perfscore{100}{32.6}} & \perfscore{83}{10.3} & \textbf{\perfscore{100}{82.1}} & \textbf{\perfscore{100}{65.2}} & \perfscore{93}{31.9} \\
    \bottomrule
  \end{tabular}%
  }
\end{table}

\paragraph{Separate-teacher results.}
\Cref{tab:opd-avg16} shows that \ours{} achieves the highest math and code macro-averages at both student scales. \ours{} achieves the best or tied-best results on $7/8$ benchmarks ($4/5$ in math) for Qwen3-1.7B and $5/8$ benchmarks ($3/5$ in math) for Qwen3-4B. On Qwen3-4B-Instruct-2507, \ours{} improves math macro-average Avg@16 over Base and vanilla OPD by $4.56$ and $2.08$ percentage points, respectively. The largest absolute gains occur on AIME24, where \ours{} reaches $65.4\%$, corresponding to relative improvements of $10.7\%$ over Base and $9.0\%$ over vanilla OPD ($+6.3$pp and $+5.4$pp, respectively). These gains over vanilla OPD suggest that a fixed teacher can provide more effective supervision when distillation follows refined trajectories. The gains over SKD may arise because \ours{} can revise earlier reasoning decisions using the full rollout as context, whereas SKD's token-level interventions leave previously accepted prefixes unchanged.

\begin{table}[!h]
  \vspace{-1em}
  \caption{Privileged self-teacher Avg@16 results (\%). The +OPSD family uses a stop-gradient teacher that shares the student backbone and is conditioned on $y^\star$. Bold marks block best.}
  \label{tab:opsd-avg16}
  \centering
  \scriptsize
  \setlength{\tabcolsep}{2.2pt}
  \resizebox{\textwidth}{!}{%
  \begin{tabular}{llcccccccc}
    \toprule
    Method & Traj. & \multicolumn{5}{c}{Math} & \multicolumn{3}{c}{Code} \\
    \cmidrule(lr){3-7}\cmidrule(lr){8-10}
    & & AIME24 & AIME25 & HMMT25 & BeyondAIME & AMOBench & HumanEval+ & MBPP+ & LiveCodeBench\\
    \midrule
    \multicolumn{10}{l}{\emph{Qwen3-4B-Instruct-2507}} \\
    \rowcolor{gray!15}
    Base & -- & \textcolor[HTML]{707070}{59.1} & \textcolor[HTML]{707070}{42.9} & \textcolor[HTML]{707070}{28.3} & \textcolor[HTML]{707070}{28.2} & \textcolor[HTML]{707070}{8.1} & \textcolor[HTML]{707070}{78.0} & \textcolor[HTML]{707070}{60.1} & \textcolor[HTML]{707070}{28.1} \\
    +SFT & $y^\star$ & \perfscore{49}{60.7} & \perfscore{40}{44.3} & \perfscore{50}{28.6} & \perfscore{40}{29.4} & \perfscore{40}{8.3} & \perfscore{40}{79.3} & \perfscore{40}{61.4} & \perfscore{40}{29.1} \\
    \rowcolor{gray!15}
    +GRPO & $y_o$ & \perfscore{83}{62.3} & \perfscore{93}{48.8} & \perfscore{91}{31.5} & \perfscore{85}{31.9} & \perfscore{98}{11.1} & \perfscore{61}{80.3} & \perfscore{77}{63.6} & \textbf{\perfscore{100}{32.3}} \\
    \arrayrulecolor{black!25}\midrule\arrayrulecolor{black}
    +OPSD & $y_o$ & \perfscore{40}{60.3} & \perfscore{93}{48.8} & \perfscore{81}{30.8} & \perfscore{93}{32.3} & \perfscore{65}{9.5} & \perfscore{79}{81.1} & \textbf{\perfscore{100}{65.0}} & \perfscore{93}{31.9} \\
    \rowcolor{gray!15}
    +OPSD (w/ Top-$K$) & $y_o$ & \perfscore{98}{63.0} & \perfscore{95}{49.0} & \perfscore{84}{31.0} & \perfscore{93}{32.3} & \perfscore{67}{9.6} & \perfscore{91}{81.7} & \perfscore{95}{64.7} & \perfscore{96}{32.1} \\
    +OPSD (w/ Clip) & $y_o$ & \perfscore{79}{62.1} & \perfscore{98}{49.2} & \perfscore{40}{27.9} & \perfscore{95}{32.4} & \perfscore{61}{9.3} & \textbf{\perfscore{100}{82.1}} & \perfscore{97}{64.8} & \perfscore{93}{31.9} \\
    \rowcolor{gray!15}
    +SKD & $y_o + y_r$ & \perfscore{61}{61.3} & \perfscore{60}{46.0} & \perfscore{70}{30.0} & \perfscore{75}{31.3} & \perfscore{61}{9.3} & \perfscore{83}{81.3} & \perfscore{97}{64.8} & \perfscore{94}{32.0} \\
    \midrule
    +\ours{} (ours) & $y_r$ & \textbf{\perfscore{100}{63.1}} & \textbf{\perfscore{100}{49.4}} & \textbf{\perfscore{100}{32.1}} & \textbf{\perfscore{100}{32.7}} & \textbf{\perfscore{100}{11.2}} & \perfscore{94}{81.8} & \perfscore{97}{64.8} & \perfscore{91}{31.8} \\
    \midrule
    \multicolumn{10}{l}{\emph{Qwen3-8B (w/ thinking)}} \\
    \rowcolor{gray!15}
    Base & -- & \textcolor[HTML]{707070}{72.3} & \textcolor[HTML]{707070}{59.7} & \textcolor[HTML]{707070}{30.5} & \textcolor[HTML]{707070}{35.6} & \textcolor[HTML]{707070}{8.9} & \textcolor[HTML]{707070}{70.3} & \textcolor[HTML]{707070}{57.6} & \textcolor[HTML]{707070}{47.5} \\
    +SFT & $y^\star$ & \perfscore{55}{73.2} & \perfscore{40}{61.9} & \perfscore{40}{32.9} & \perfscore{40}{38.5} & \perfscore{40}{11.3} & \perfscore{40}{73.1} & \perfscore{40}{59.8} & \perfscore{40}{49.2} \\
    \rowcolor{gray!15}
    +GRPO & $y_o$ & \perfscore{88}{75.6} & \perfscore{94}{68.5} & \perfscore{87}{41.9} & \textbf{\perfscore{100}{42.9}} & \perfscore{86}{15.9} & \perfscore{89}{78.5} & \perfscore{98}{62.5} & \perfscore{82}{52.1} \\
    \arrayrulecolor{black!25}\midrule\arrayrulecolor{black}
    +OPSD & $y_o$ & \perfscore{77}{74.8} & \perfscore{69}{65.4} & \perfscore{78}{40.3} & \perfscore{55}{39.6} & \perfscore{84}{15.7} & \perfscore{98}{79.5} & \perfscore{91}{62.2} & \perfscore{85}{52.3} \\
    \rowcolor{gray!15}
    +OPSD (w/ Top-$K$) & $y_o$ & \perfscore{88}{75.6} & \perfscore{94}{68.5} & \perfscore{99}{44.4} & \perfscore{86}{41.9} & \perfscore{79}{15.2} & \perfscore{91}{78.7} & \perfscore{91}{62.2} & \perfscore{82}{52.1} \\
    +OPSD (w/ Clip) & $y_o$ & \textbf{\perfscore{100}{76.5}} & \perfscore{93}{68.3} & \perfscore{90}{42.5} & \perfscore{63}{40.2} & \perfscore{86}{15.9} & \perfscore{92}{78.8} & \perfscore{87}{62.0} & \perfscore{93}{52.8} \\
    \rowcolor{gray!15}
    +SKD & $y_o + y_r$ & \perfscore{40}{72.1} & \perfscore{79}{66.7} & \perfscore{89}{42.3} & \perfscore{56}{39.7} & \perfscore{83}{15.6} & \perfscore{99}{79.6} & \perfscore{94}{62.3} & \perfscore{77}{51.7} \\
    \midrule
    +\ours{} (ours) & $y_r$ & \textbf{\perfscore{100}{76.5}} & \textbf{\perfscore{100}{69.2}} & \textbf{\perfscore{100}{44.5}} & \perfscore{99}{42.8} & \textbf{\perfscore{100}{17.3}} & \textbf{\perfscore{100}{79.7}} & \textbf{\perfscore{100}{62.6}} & \textbf{\perfscore{100}{53.3}} \\
    \bottomrule
  \end{tabular}%
  }
\end{table}

\paragraph{Privileged self-teacher results.}
\Cref{tab:opsd-avg16} shows that \ours{} achieves the highest math macro-average at both student scales and the highest code macro-average on Qwen3-8B. \ours{} achieves the best or tied-best results on $5/8$ benchmarks ($5/5$ in math) for Qwen3-4B and $7/8$ benchmarks ($4/5$ in math) for Qwen3-8B. On Qwen3-8B, \ours{} improves math macro-average Avg@16 over Base and vanilla OPSD by $8.66$ and $2.90$ percentage points, respectively. The largest absolute gains occur on HMMT25, where \ours{} reaches $44.5\%$, corresponding to relative improvements of $45.9\%$ over Base and $10.4\%$ over vanilla OPSD ($+14.0$pp and $+4.2$pp, respectively). These trends align with the separate-teacher results, extending the benefits of trajectory refinement to privileged self-distillation.

\subsection{\ours{} Enhances Exploration}
\label{subsec:exploration}

Beyond averaged completion accuracy, we evaluate whether \ours{} improves the student's ability to discover correct solutions as the test-time sampling budget increases. We measure this aspect of exploration with Pass@$k$, the probability of obtaining at least one correct completion within $k$ samples, averaged across questions. Full Pass@16 comparisons across math and code under both teacher regimes are provided in \cref{app:pass16-results}.

\begin{wrapfigure}{r}{0.4\textwidth}
  \centering
  \vspace{-3em}
  \includegraphics[width=\linewidth,trim=368bp 20bp 0bp 16bp,clip]{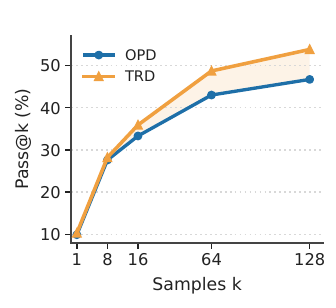}
  \vspace{-2em}
  \caption{Pass@$k$ on AMOBench for Qwen3-4B-Instruct-2507 under the separate-teacher regime.}
  \label{fig:exploration-passk}
\end{wrapfigure}

\cref{fig:exploration-passk} compares vanilla OPD trained on $y_o$ with \ours{} trained on $y_r$. The gain is modest at $k{=}1$, but widens as the sampling budget increases: at $k{=}128$, \ours{} reaches $53.8\%$ compared with $46.7\%$ for vanilla OPD, an improvement of $7.1$ percentage points ($15.2\%$ relative).

These results show that \ours{} broadens solution coverage as the sampling budget grows. The trajectory analyses in \cref{exp:trajectory} provide complementary evidence: refinement improves verifier accuracy and produces shorter solutions even for initially correct training rollouts. At evaluation, vanilla OPD and \ours{} exhibit similar correct-rollout length distributions, so the coverage gain is accompanied by comparable lengths of successful solutions.

\WFclear
\subsection{Validating Trajectory Construction}
\label{subsec:objective-validation}

We test whether optimizing \cref{eq:prefix-refinement-objective} by selecting successful on-policy trajectories provides an effective surrogate for mitigating prefix failure and improving downstream performance. As discussed in \cref{sec:method}, rejection sampling offers one way to pursue this objective while respecting the support constraint. We implement this strategy under OPSD through Best-of-$N$ filtering, drawing $N$ student rollouts per problem and retaining verifier-passing trajectories.

\Needspace{0.4\textheight}
\begin{wrapfigure}{r}{0.44\textwidth}
  \vspace{-2.0em}
  \centering
  \includegraphics[width=\linewidth]{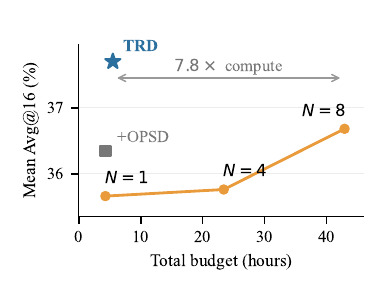}
  \caption{Full-corpus Best-of-$N$ scaling under OPSD with Qwen3-4B-Instruct-2507 on math tasks.}
  \label{fig:eq5-best-of-n-scaling}
\end{wrapfigure}

\Cref{fig:eq5-best-of-n-scaling} compares Best-of-$N$ with \ours{} on the full Qwen3-4B-Instruct-2507 training corpus, allowing sampling cost to grow with $N$. Increasing $N$ improves downstream performance, supporting the usefulness of \cref{eq:prefix-refinement-objective} as a surrogate for mitigating prefix failure. However, finding successful trajectories through repeated sampling incurs substantial compute costs: even at $7.8\times$ the total wall-clock time of \ours{}, Best-of-8 does not match its performance. By using privileged guidance to refine sampled trajectories, \ours{} achieves stronger results with substantially less compute.

Detailed experimental results are presented in \cref{app:objective-validation}, including budget-matched Best-of-$N$ comparisons, which also favor \ours{} overall (\cref{tab:matched-best-of-n}). For further analysis of \cref{eq:prefix-refinement-objective}, \cref{tab:trajectory-source-comparison} shows that alternative supervision trajectories that may violate the support constraint underperform TRD overall.

\section{Conclusion}
\label{sec:conclusion}

We identify \emph{prefix failure} as a structural limitation of on-policy (self)-distillation paradigms, where the per-token KL evaluated along the student's frozen rollout induces a bimodal teacher mixture and a fragmented gradient that loss-level fixes leave structurally intact. To address it, we propose \ours{}, a trajectory-level refinement that draws a refined trajectory $y_r$ under teacher guidance and supervises the per-token KL along $y_r$, recovering the ideal supervision-pair structure while remaining on-policy. Under OPSD, Qwen3-8B with \ours{} solves $9/23$ AMOBench questions left unsolved by the base model in $16$ attempts, nearly doubling the strongest dense-KL baseline's Pass@16 on this subset (\cref{app:test-rollout}). \ours{} also outperforms Best-of-8, which requires $7.8\times$ total wall-clock time.

\paragraph{Limitations.}
First, \ours{} requires one extra sampling budget to construct $y_r$. This overhead is partially offset by faster KL training on shorter refined trajectories; on Qwen3-8B, the total wall-clock nearly matches the dense-KL baselines (\cref{app:hardware}). Second, \ours{} relies on the teacher's ability to guide refinement in a way that mitigates prefix failure while keeping the refined trajectories close to the student's on-policy distribution. This limitation is less severe with a stronger teacher that can in principle optimize \cref{eq:prefix-refinement-objective} with promise.



{
\bibliographystyle{iclr2027_conference}
\bibliography{references}
}

\raggedbottom
\newpage
\appendix

\begin{center}
\vspace*{0.5em}
{\large\bfseries Appendix: \oursfull{}}
\vspace*{0.5em}
\end{center}
\section{Derivation of the OPD Policy Gradient}
\label[appendix]{app:opd-derivation}

We derive the exact sequence-level reverse-KL gradient, then distinguish the immediate-token surrogate in \cref{eq:opsd_grad}. The derivation parallels~\citet{yang2026gopd}. Throughout, the teacher $\pi_T$ is held fixed during differentiation, using stop-gradient in OPSD, and $\delta_t \!:=\! \log \pi_\theta(y_t\mid x,y_{<t}) - \log \pi_T(y_t\mid x,y_{<t})$.

\paragraph{Step 1 (KL as a log-ratio expectation).}
The unnormalized sequence-level reverse KL $\mathcal{J}_{\mathrm{seq}}(\theta) = \mathbb{E}_{x\sim\mathcal{D}}[D_{\mathrm{KL}}(\pi_\theta(\cdot\mid x)\,\|\,\pi_T(\cdot\mid x))]$ expands to
\begin{equation*}
    \mathcal{J}_{\mathrm{seq}}(\theta)
    \;=\;
    \mathbb{E}_{x\sim\mathcal{D},\,y\sim \pi_\theta(\cdot\mid x)}
    \!\big[\log \pi_\theta(y\mid x) - \log \pi_T(y\mid x)\big],
\end{equation*}
where $\theta$ enters through both the sampling distribution and the integrand. This objective is distinct from the length-normalized token losses in \cref{eq:opd_loss,eq:opsd_loss}.

\paragraph{Step 2 (product rule and score-function trick).}
Differentiating under the expectation gives
\begin{equation*}
\begin{aligned}
    \nabla_{\theta}\mathcal{J}_{\mathrm{seq}}(\theta)
    \;=\;
    \mathbb{E}_{x}\!\bigg[
    &\sum_y \big(\nabla_{\theta} \pi_\theta(y\mid x)\big)\big(\log \pi_\theta(y\mid x) - \log \pi_T(y\mid x)\big) \\
    &+\;
    \sum_y \pi_\theta(y\mid x)\,\nabla_{\theta}\log \pi_\theta(y\mid x)
    \bigg].
\end{aligned}
\end{equation*}
The second sum vanishes because
\begin{equation*}
    \sum_y \pi_\theta(y\mid x)\,\nabla_{\theta}\log \pi_\theta(y\mid x)
    \;=\;
    \sum_y \nabla_{\theta} \pi_\theta(y\mid x)
    \;=\;
    \nabla_{\theta} 1 \;=\; 0.
\end{equation*}
Using $\nabla_{\theta} \pi_\theta = \pi_\theta\,\nabla_{\theta}\log \pi_\theta$, the remaining term becomes
\begin{equation}
\label{eq:opd_grad_seq_app}
    \nabla_{\theta}\mathcal{J}_{\mathrm{seq}}(\theta)
    \;=\;
    \mathbb{E}_{x\sim\mathcal{D},\,y\sim \pi_\theta(\cdot\mid x)}
    \!\big[
    \big(\log \pi_\theta(y\mid x) - \log \pi_T(y\mid x)\big)
    \nabla_{\theta}\log \pi_\theta(y\mid x)
    \big].
\end{equation}

\paragraph{Step 3 (autoregressive decomposition).}
Factor $\log \pi_\theta(y\mid x)\!=\!\sum_{t} \log \pi_\theta(y_{t}\mid x,y_{<t})$ and similarly for $\pi_T$. \Cref{eq:opd_grad_seq_app} expands to
\begin{equation*}
    \mathbb{E}_{x,y}\!\left[
    \sum_{t=1}^{|y|}\sum_{t'=1}^{|y|}
    \underbrace{\big(\log \pi_\theta(y_{t'}\mid x,y_{<t'}) - \log \pi_T(y_{t'}\mid x,y_{<t'})\big)}_{\delta_{t'}}\,
    \nabla_{\theta}\log \pi_\theta(y_t\mid x,y_{<t})
    \right].
\end{equation*}

\paragraph{Step 4 (causality, past log-ratios do not contribute).}
For any $t' < t$, $\delta_{t'}$ is measurable with respect to $(x,y_{<t})$, and conditioning on this prefix yields
\begin{equation*}
    \mathbb{E}_{y_t\sim \pi_\theta(\cdot\mid x,y_{<t})}\!\big[\nabla_{\theta}\log \pi_\theta(y_t\mid x,y_{<t})\big]
    \;=\;
    \sum_{y_t} \nabla_{\theta} \pi_\theta(y_t\mid x,y_{<t})
    \;=\;
    \nabla_{\theta} 1 \;=\; 0,
\end{equation*}
so all cross terms with $t' < t$ vanish in expectation.

\paragraph{Step 5 (exact gradient and per-token surrogate).}
Retaining $t' \ge t$ gives the exact sequence-level gradient,
\begin{equation*}
    \nabla_{\theta}\mathcal{J}_{\mathrm{seq}}(\theta)
    \;=\;
    \mathbb{E}_{x,y}\!\left[
    \sum_{t=1}^{|y|}
    \Bigg(\sum_{t'=t}^{|y|} \delta_{t'}\Bigg)\,
    \nabla_{\theta}\log \pi_\theta(y_t\mid x,y_{<t})
    \right].
\end{equation*}
For gradient descent, $-\!\sum_{t' \ge t}\!\delta_{t'}$ acts as a return-to-go for token $y_t$. Discarding future log-ratios~\citep{agarwal2024policy,lu2025onpolicy} gives the immediate-token surrogate used in \cref{eq:opsd_grad},
\begin{equation*}
    g_{\mathrm{tok}}(\theta)
    \;:=\;
    \mathbb{E}_{x,y}\!\left[
    \sum_{t=1}^{|y|}
    \big(\log \pi_\theta(y_t\mid x,y_{<t}) - \log \pi_T(y_t\mid x,y_{<t})\big)\,
    \nabla_{\theta}\log \pi_\theta(y_t\mid x,y_{<t})
    \right].
\end{equation*}
This surrogate omits future-return terms and is not the exact gradient of $\mathcal{J}_{\mathrm{seq}}$. In \cref{eq:opd_loss,eq:opsd_loss}, implementations instead hold sampled trajectories fixed, evaluate full-vocabulary token KL, and apply the stated $1/|y|$ normalization. OPSD additionally conditions the stop-gradient teacher on $y^\star$.

\section{Additional Experiments}
\label[appendix]{app:add-exp}

This section provides supporting analyses for the main experiments. \Cref{subsec:exp_prefix} examines training diagnostics related to prefix failure, and \cref{app:pass16-results} reports Pass@16 under both teacher regimes. \Cref{app:objective-validation} examines alternative trajectory sources and Best-of-$N$ filtering to validate the trajectory-construction objective and its compute trade-offs. \Cref{subsec:trajectory-analysis} examines training-trajectory quality and length in the privileged self-teacher regime, together with correct evaluation-rollout lengths in the separate-teacher regime. \Cref{subsec:opd-opsd-trd} compares the separate teacher and privileged self-teacher for \ours{} on Qwen3-4B-Instruct-2507. \Cref{app:test-rollout} reports AMOBench test-rollout coverage by base-model difficulty for Qwen3-8B under OPSD.

\subsection{Experimental Validation of Prefix Failure}
\label[appendix]{subsec:exp_prefix}

\begin{figure}[!htbp]
\centering
\includegraphics[width=\linewidth]{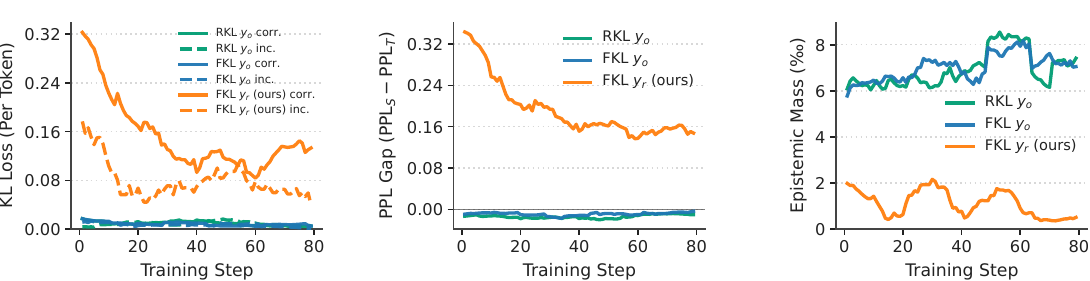}
\caption{Training diagnostics in the privileged self-teacher setting: standard OPSD on $y_o$ with forward or reverse KL, and \ours{} on $y_r$ with forward KL. \textbf{Left}: Token-weighted mean KL, grouped by the original stage-1 rollout's verifier outcome. \textbf{Middle}: Teacher-student perplexity gap on the supervised response. \textbf{Right}: Mean teacher probability mass on a fixed epistemic-token set. Metric definitions and normalization are given in \cref{app:training-metrics}.}
\label{fig:prefix-failure-evidence}
\end{figure}

We examine three training diagnostics related to the prefix-failure analysis in \cref{sec:prefix_failure}: per-token KL, the teacher-student perplexity gap, and epistemic-token mass (\cref{fig:prefix-failure-evidence}). Standard OPSD is evaluated on raw student rollouts $y_o$ under forward and reverse KL; \ours{} is evaluated on refined trajectories $y_r$ under forward KL. These measurements characterize supervision on the respective trajectories. They do not directly locate failed prefixes or establish the causal contribution of prefix correction to downstream gains.

\paragraph{Per-token Distributional Discrepancy (left).} We group responses by the original stage-1 rollout's verifier outcome and compute token-weighted mean KL values $D^{\text{correct}}$ and $D^{\text{incorrect}}$. Each mean divides the summed token-level KL by the number of unmasked response tokens, so it is already normalized for token count (\cref{app:training-metrics}). The same outcome labels are used for the paired $y_r$ responses. On $y_o$, both outcome groups exhibit small average KL, including the group whose original responses fail verification. Within the forward-KL runs, $y_r$ exhibits a larger average teacher-student discrepancy than $y_o$. This comparison is descriptive: token-count normalization does not remove differences in trajectory content or prefix distributions, and forward and reverse KL are distinct quantities. A smaller mean KL does not imply an uninformative gradient, nor does a larger value establish more effective supervision. The observation on incorrect $y_o$ shows that local distributional agreement can coexist with an incorrect final answer.

\paragraph{Teacher-student Perplexity Gap (middle).} For each run, we compute both perplexities by exponentiating the token-weighted mean negative log-likelihood on the same supervised response tokens. The gap $\mathrm{PPL}_S-\mathrm{PPL}_T$ remains small on $y_o$, whereas $y_r$ initially exhibits a larger positive gap that decreases during training. This describes the models' relative likelihoods of the observed responses. Since the response distributions differ across runs, the gap alone does not measure the usefulness of privileged information or establish that a correction path has been learned.

\paragraph{Epistemic-token Mass (right).} On $y_o$, the teacher assigns approximately $6$ to $8$\,\textperthousand{} of its average next-token probability mass to a fixed token set constructed from $16$ correction-related phrases (\cref{app:training-metrics}); on $y_r$, this quantity falls below $2$\,\textperthousand{}. The higher mass on raw trajectories is consistent with more frequent correction-related cues under failed-prefix conditioning, as discussed in \cref{subsec:perfect_teacher}. However, these tokens also occur in valid reasoning, and their aggregate mass neither identifies failed prefixes nor demonstrates repeated correction onsets at specific positions. We therefore treat this pattern as a supporting diagnostic of changed conditioning, rather than direct evidence of gradient fragmentation.

\subsection{Pass@16 Results}
\label[appendix]{app:pass16-results}

This subsection provides the full Pass@16 results accompanying the exploration analysis in \cref{subsec:exploration}. Pass@16 measures the fraction of questions with at least one correct completion among $16$ samples, complementing the Avg@16 results in \cref{exp:main}. \Cref{tab:opd-pass16,tab:opsd-pass16} report this coverage metric for the same methods and model configurations as \cref{tab:opd-avg16,tab:opsd-avg16}, using the evaluation protocol in \cref{exp:setup}.

\begin{table}[!h]
  \caption{Separate-teacher Pass@16 results (\%). The +OPD family uses a frozen Qwen3-8B teacher without access to $y^\star$. Bold marks block best.}
  \label{tab:opd-pass16}
  \centering
  \scriptsize
  \setlength{\tabcolsep}{2.2pt}
  \resizebox{\textwidth}{!}{%
  \begin{tabular}{llcccccccc}
    \toprule
    Method & Traj. & \multicolumn{5}{c}{Math} & \multicolumn{3}{c}{Code} \\
    \cmidrule(lr){3-7}\cmidrule(lr){8-10}
    & & AIME24 & AIME25 & HMMT25 & BeyondAIME & AMOBench & HumanEval+ & MBPP+ & LiveCodeBench\\
    \midrule
    \multicolumn{10}{l}{\emph{Qwen3-1.7B (w/ thinking)}} \\
    \rowcolor{gray!15}
    Base & -- & \textcolor[HTML]{707070}{63.3} & \textbf{\textcolor[HTML]{707070}{66.7}} & \textcolor[HTML]{707070}{46.7} & \textcolor[HTML]{707070}{42.0} & \textcolor[HTML]{707070}{12.8} & \textcolor[HTML]{707070}{68.2} & \textcolor[HTML]{707070}{55.1} & \textcolor[HTML]{707070}{40.3} \\
    +SFT & $y^\star$ & \perfscore{40}{60.0} & \perfscore{80}{63.3} & \perfscore{40}{43.3} & \perfscore{70}{39.0} & \perfscore{55}{10.3} & \perfscore{40}{65.8} & \perfscore{40}{50.8} & \perfscore{40}{41.2} \\
    \rowcolor{gray!15}
    +GRPO & $y_o$ & \perfscore{90}{76.6} & \perfscore{80}{63.3} & \textbf{\perfscore{100}{53.3}} & \perfscore{89}{44.0} & \perfscore{70}{12.8} & \perfscore{91}{76.2} & \perfscore{90}{62.7} & \textbf{\perfscore{100}{48.3}} \\
    \arrayrulecolor{black!25}\midrule\arrayrulecolor{black}
    +OPD & $y_o$ & \perfscore{90}{76.7} & \perfscore{80}{63.3} & \textbf{\perfscore{100}{53.3}} & \perfscore{89}{44.0} & \perfscore{70}{12.8} & \perfscore{67}{71.3} & \perfscore{71}{58.2} & \perfscore{42}{41.4} \\
    \rowcolor{gray!15}
    +OPD (w/ Top-$K$) & $y_o$ & \perfscore{90}{76.7} & \perfscore{60}{60.0} & \perfscore{60}{46.7} & \perfscore{89}{44.0} & \perfscore{85}{15.4} & \textbf{\perfscore{100}{78.0}} & \perfscore{88}{62.2} & \perfscore{86}{46.6} \\
    +OPD (w/ Clip) & $y_o$ & \textbf{\perfscore{100}{80.0}} & \perfscore{40}{56.7} & \perfscore{80}{50.0} & \textbf{\perfscore{100}{47.0}} & \perfscore{55}{10.3} & \perfscore{97}{77.4} & \perfscore{86}{61.9} & \perfscore{91}{47.2} \\
    \rowcolor{gray!15}
    +SKD & $y_o + y_r$ & \perfscore{80}{73.3} & \perfscore{40}{56.7} & \perfscore{40}{43.3} & \perfscore{40}{31.0} & \perfscore{40}{7.7} & \perfscore{67}{71.3} & \perfscore{66}{57.1} & \perfscore{90}{47.1} \\
    \midrule
    +\ours{} (ours) & $y_r$ & \textbf{\perfscore{100}{80.0}} & \textbf{\perfscore{100}{66.7}} & \textbf{\perfscore{100}{53.3}} & \perfscore{92}{45.0} & \textbf{\perfscore{100}{17.9}} & \textbf{\perfscore{100}{78.0}} & \textbf{\perfscore{100}{65.2}} & \perfscore{87}{46.8} \\
    \midrule
    \multicolumn{10}{l}{\emph{Qwen3-4B-Instruct-2507}} \\
    \rowcolor{gray!15}
    Base & -- & \textcolor[HTML]{707070}{73.3} & \textcolor[HTML]{707070}{66.7} & \textcolor[HTML]{707070}{43.3} & \textcolor[HTML]{707070}{54.0} & \textcolor[HTML]{707070}{23.1} & \textcolor[HTML]{707070}{79.2} & \textcolor[HTML]{707070}{68.7} & \textcolor[HTML]{707070}{51.2} \\
    +SFT & $y^\star$ & \perfscore{40}{73.3} & \perfscore{40}{66.7} & \perfscore{40}{46.6} & \perfscore{40}{55.0} & \perfscore{40}{22.6} & \perfscore{40}{81.1} & \perfscore{40}{69.7} & \perfscore{40}{52.0} \\
    \rowcolor{gray!15}
    +GRPO & $y_o$ & \perfscore{85}{83.3} & \perfscore{80}{73.3} & \perfscore{80}{53.3} & \textbf{\perfscore{100}{62.0}} & \perfscore{65}{28.2} & \perfscore{95}{87.8} & \perfscore{77}{72.2} & \perfscore{84}{54.5} \\
    \arrayrulecolor{black!25}\midrule\arrayrulecolor{black}
    +OPD & $y_o$ & \perfscore{70}{80.0} & \perfscore{80}{73.3} & \perfscore{60}{50.0} & \perfscore{91}{61.0} & \perfscore{88}{33.3} & \perfscore{90}{87.2} & \perfscore{82}{72.6} & \perfscore{63}{53.3} \\
    \rowcolor{gray!15}
    +OPD (w/ Top-$K$) & $y_o$ & \perfscore{70}{80.0} & \perfscore{60}{70.0} & \perfscore{80}{53.3} & \perfscore{57}{57.0} & \perfscore{65}{28.2} & \perfscore{90}{87.2} & \perfscore{77}{72.2} & \perfscore{84}{54.5} \\
    +OPD (w/ Clip) & $y_o$ & \perfscore{85}{83.3} & \perfscore{80}{73.3} & \perfscore{80}{53.3} & \perfscore{91}{61.0} & \perfscore{54}{25.6} & \textbf{\perfscore{100}{88.4}} & \perfscore{93}{73.3} & \perfscore{77}{54.1} \\
    \rowcolor{gray!15}
    +SKD & $y_o + y_r$ & \textbf{\perfscore{100}{86.7}} & \perfscore{80}{73.3} & \textbf{\perfscore{100}{56.7}} & \textbf{\perfscore{100}{62.0}} & \perfscore{54}{25.6} & \perfscore{90}{87.2} & \perfscore{96}{73.5} & \textbf{\perfscore{100}{55.4}} \\
    \midrule
    +\ours{} (ours) & $y_r$ & \perfscore{85}{83.3} & \textbf{\perfscore{100}{76.7}} & \perfscore{60}{50.0} & \textbf{\perfscore{100}{62.0}} & \textbf{\perfscore{100}{35.9}} & \textbf{\perfscore{100}{88.4}} & \textbf{\perfscore{100}{73.8}} & \perfscore{77}{54.1} \\
    \bottomrule
  \end{tabular}%
  }
\end{table}

\paragraph{Separate-teacher results.}
\Cref{tab:opd-pass16} reports results with the frozen Qwen3-8B teacher. On Qwen3-1.7B, \ours{} ties the best AIME24, AIME25, and HMMT25 results and leads AMOBench at $17.9\%$, while +OPD (w/ Clip) leads BeyondAIME. On Qwen3-4B-Instruct-2507, \ours{} leads AIME25 and AMOBench and ties the best BeyondAIME result, while SKD leads AIME24 and HMMT25. Its AMOBench score rises from Base's $23.1\%$ to $35.9\%$, a gain of $12.8$. On code, \ours{} matches the best HumanEval+ value and leads MBPP+ at both student scales. GRPO and SKD give the strongest LiveCodeBench results on Qwen3-1.7B and Qwen3-4B, respectively. These results show that refinement improves coverage across most benchmarks, although other methods remain stronger on some math and code tasks.

\begin{table}[!h]
  \caption{Privileged self-teacher Pass@16 results (\%). The +OPSD family uses a stop-gradient teacher that shares the student backbone and is conditioned on $y^\star$. Bold marks block best.}
  \label{tab:opsd-pass16}
  \centering
  \scriptsize
  \setlength{\tabcolsep}{2.2pt}
  \resizebox{\textwidth}{!}{%
  \begin{tabular}{llcccccccc}
    \toprule
    Method & Traj. & \multicolumn{5}{c}{Math} & \multicolumn{3}{c}{Code} \\
    \cmidrule(lr){3-7}\cmidrule(lr){8-10}
    & & AIME24 & AIME25 & HMMT25 & BeyondAIME & AMOBench & HumanEval+ & MBPP+ & LiveCodeBench\\
    \midrule
    \multicolumn{10}{l}{\emph{Qwen3-4B-Instruct-2507}} \\
    \rowcolor{gray!15}
    Base & -- & \textcolor[HTML]{707070}{73.3} & \textcolor[HTML]{707070}{66.7} & \textcolor[HTML]{707070}{43.3} & \textcolor[HTML]{707070}{54.0} & \textcolor[HTML]{707070}{23.1} & \textcolor[HTML]{707070}{79.2} & \textcolor[HTML]{707070}{68.7} & \textcolor[HTML]{707070}{51.2} \\
    +SFT & $y^\star$ & \perfscore{40}{73.3} & \perfscore{40}{66.7} & \perfscore{40}{46.6} & \perfscore{40}{55.0} & \perfscore{40}{22.6} & \perfscore{40}{81.1} & \perfscore{49}{69.7} & \perfscore{77}{52.0} \\
    \rowcolor{gray!15}
    +GRPO & $y_o$ & \perfscore{85}{83.3} & \perfscore{70}{73.3} & \perfscore{80}{53.3} & \perfscore{87}{62.0} & \perfscore{70}{28.2} & \perfscore{95}{87.8} & \perfscore{83}{72.2} & \perfscore{93}{54.5} \\
    \arrayrulecolor{black!25}\midrule\arrayrulecolor{black}
    +OPSD & $y_o$ & \textbf{\perfscore{100}{86.7}} & \perfscore{85}{76.7} & \perfscore{80}{53.3} & \perfscore{60}{58.0} & \perfscore{97}{33.3} & \perfscore{90}{87.2} & \textbf{\perfscore{100}{73.5}} & \perfscore{98}{55.3} \\
    \rowcolor{gray!15}
    +OPSD (w/ Top-$K$) & $y_o$ & \textbf{\perfscore{100}{86.7}} & \perfscore{85}{76.7} & \textbf{\perfscore{100}{56.7}} & \perfscore{53}{57.0} & \perfscore{56}{25.6} & \perfscore{95}{87.8} & \perfscore{91}{72.8} & \textbf{\perfscore{100}{55.6}} \\
    +OPSD (w/ Clip) & $y_o$ & \textbf{\perfscore{100}{86.7}} & \perfscore{85}{76.7} & \perfscore{41}{46.7} & \perfscore{87}{62.0} & \perfscore{70}{28.2} & \perfscore{90}{87.2} & \perfscore{91}{72.8} & \perfscore{96}{54.9} \\
    \rowcolor{gray!15}
    +SKD & $y_o + y_r$ & \perfscore{70}{80.0} & \perfscore{40}{66.7} & \perfscore{80}{53.3} & \perfscore{53}{57.0} & \perfscore{70}{28.2} & \perfscore{75}{85.4} & \perfscore{40}{69.0} & \perfscore{40}{46.1} \\
    \midrule
    +\ours{} (ours) & $y_r$ & \textbf{\perfscore{100}{86.7}} & \textbf{\perfscore{100}{80.0}} & \textbf{\perfscore{100}{56.7}} & \textbf{\perfscore{100}{64.0}} & \textbf{\perfscore{100}{33.8}} & \textbf{\perfscore{100}{88.4}} & \perfscore{97}{73.3} & \perfscore{97}{55.2} \\
    \midrule
    \multicolumn{10}{l}{\emph{Qwen3-8B (w/ thinking)}} \\
    \rowcolor{gray!15}
    Base & -- & \textcolor[HTML]{707070}{83.3} & \textcolor[HTML]{707070}{76.6} & \textcolor[HTML]{707070}{66.7} & \textcolor[HTML]{707070}{62.0} & \textcolor[HTML]{707070}{41.0} & \textcolor[HTML]{707070}{84.5} & \textcolor[HTML]{707070}{69.3} & \textcolor[HTML]{707070}{64.1} \\
    +SFT & $y^\star$ & \perfscore{42}{86.7} & \perfscore{40}{80.0} & \perfscore{61}{70.0} & \perfscore{62}{64.0} & \perfscore{60}{46.2} & \perfscore{40}{82.3} & \perfscore{40}{68.7} & \perfscore{40}{63.4} \\
    \rowcolor{gray!15}
    +GRPO & $y_o$ & \textbf{\perfscore{100}{90.0}} & \perfscore{70}{83.3} & \perfscore{61}{70.0} & \perfscore{78}{66.0} & \perfscore{73}{51.3} & \textbf{\perfscore{100}{89.0}} & \textbf{\perfscore{100}{72.8}} & \perfscore{91}{68.2} \\
    \arrayrulecolor{black!25}\midrule\arrayrulecolor{black}
    +OPSD & $y_o$ & \perfscore{42}{86.7} & \perfscore{70}{83.3} & \perfscore{81}{73.3} & \perfscore{92}{68.0} & \perfscore{73}{51.3} & \perfscore{96}{88.6} & \perfscore{84}{71.7} & \perfscore{93}{68.4} \\
    \rowcolor{gray!15}
    +OPSD (w/ Top-$K$) & $y_o$ & \perfscore{42}{86.7} & \textbf{\perfscore{100}{86.7}} & \perfscore{40}{66.7} & \textbf{\perfscore{100}{69.0}} & \perfscore{40}{38.5} & \perfscore{95}{88.4} & \textbf{\perfscore{100}{72.8}} & \perfscore{91}{68.2} \\
    +OPSD (w/ Clip) & $y_o$ & \textbf{\perfscore{100}{90.0}} & \textbf{\perfscore{100}{86.7}} & \perfscore{61}{70.0} & \perfscore{40}{61.0} & \perfscore{53}{43.6} & \perfscore{95}{88.4} & \perfscore{77}{71.2} & \perfscore{95}{68.6} \\
    \rowcolor{gray!15}
    +SKD & $y_o + y_r$ & \perfscore{40}{86.6} & \perfscore{70}{83.3} & \perfscore{81}{73.3} & \perfscore{70}{65.0} & \perfscore{61}{46.4} & \perfscore{89}{87.8} & \perfscore{96}{72.5} & \perfscore{82}{67.4} \\
    \midrule
    +\ours{} (ours) & $y_r$ & \textbf{\perfscore{100}{90.0}} & \textbf{\perfscore{100}{86.7}} & \textbf{\perfscore{100}{76.3}} & \perfscore{92}{68.0} & \textbf{\perfscore{100}{61.5}} & \perfscore{95}{88.4} & \perfscore{96}{72.5} & \textbf{\perfscore{100}{69.1}} \\
    \bottomrule
  \end{tabular}%
  }
\end{table}

\paragraph{Privileged self-teacher results.}
\Cref{tab:opsd-pass16} reports results with the reference-conditioned self-teacher. On math, \ours{} is best or tied-best in nine of the ten model and benchmark settings. On Qwen3-4B-Instruct-2507, it leads AIME25, BeyondAIME, and AMOBench and ties the best AIME24 and HMMT25 values. On Qwen3-8B, it leads HMMT25 and AMOBench and ties the best AIME24 and AIME25 results; +OPSD (w/ Top-$K$) leads BeyondAIME. On AMOBench, \ours{} reaches $61.5\%$, compared with $51.3\%$ for both GRPO and +OPSD, the strongest baselines at this scale. On code, \ours{} leads Qwen3-4B HumanEval+ and Qwen3-8B LiveCodeBench. +OPSD leads Qwen3-4B MBPP+, and +OPSD (w/ Top-$K$) leads Qwen3-4B LiveCodeBench. On Qwen3-8B, GRPO leads HumanEval+ and ties +OPSD (w/ Top-$K$) on MBPP+. Coverage gains are therefore strongest on math, with more varied rankings across code benchmarks.

\subsection{Additional Validation of the Trajectory-Construction Objective}
\label[appendix]{app:objective-validation}

We expand the validation in \cref{subsec:objective-validation} along two dimensions of \cref{eq:prefix-refinement-objective}: the role of the support constraint and the benefit of selecting successful on-policy trajectories. All comparisons use OPSD, with privileged self-teacher supervision. The trajectory-source and budget-matched Best-of-$N$ comparisons report Avg@16 and Pass@16 on five math benchmarks for Qwen3-4B-Instruct-2507 and Qwen3-8B; the full-corpus scaling experiment in \cref{fig:eq5-best-of-n-scaling} uses Qwen3-4B-Instruct-2507. Together, these experiments assess the objective's practical value as a surrogate for mitigating prefix failure.

\begin{table}[t]
  \caption{Matched-budget trajectory-source comparison on five math benchmarks (\%). $y_t$ is generated by Qwen3-14B. Bold indicates the best or tied-best result in each model block.}
  \label{tab:trajectory-source-comparison}
  \centering
  \scriptsize
  \setlength{\tabcolsep}{1.8pt}
  \resizebox{\textwidth}{!}{%
  \begin{tabular}{lc*{10}{c}}
    \toprule
    & & \multicolumn{5}{c}{Avg@16} & \multicolumn{5}{c}{Pass@16} \\
    \cmidrule(lr){3-7}\cmidrule(lr){8-12}
    Method & Traj. & AIME24 & AIME25 & HMMT25 & BeyondAIME & AMOBench & AIME24 & AIME25 & HMMT25 & BeyondAIME & AMOBench \\
    \midrule
    \multicolumn{12}{l}{\textit{Qwen3-4B-Instruct-2507}} \\
    \rowcolor{gray!15}
    Base & -- & \textcolor[HTML]{707070}{59.1} & \textcolor[HTML]{707070}{42.9} & \textcolor[HTML]{707070}{28.3} & \textcolor[HTML]{707070}{28.2} & \textcolor[HTML]{707070}{8.1} & \textcolor[HTML]{707070}{73.3} & \textcolor[HTML]{707070}{66.7} & \textcolor[HTML]{707070}{43.3} & \textcolor[HTML]{707070}{54.0} & \textcolor[HTML]{707070}{23.1} \\
    +OPSD & $y_o$ & \perfscore{40}{60.3} & \perfscore{89}{48.8} & \perfscore{59}{30.8} & \perfscore{40}{32.3} & \perfscore{40}{9.5} & \textbf{\perfscore{100}{86.7}} & \perfscore{40}{76.7} & \perfscore{80}{53.3} & \perfscore{40}{58.0} & \perfscore{90}{33.3} \\
    \rowcolor{gray!15}
    +OPSD & $y_t$ & \perfscore{74}{61.9} & \perfscore{40}{46.2} & \perfscore{40}{30.2} & \perfscore{60}{32.5} & \perfscore{51}{9.8} & \perfscore{40}{83.3} & \perfscore{40}{76.7} & \textbf{\perfscore{100}{56.7}} & \perfscore{70}{61.0} & \perfscore{40}{30.8} \\
    +OPSD & $y^\star$ & \perfscore{87}{62.5} & \perfscore{93}{49.0} & \perfscore{65}{31.0} & \textbf{\perfscore{100}{32.9}} & \perfscore{44}{9.6} & \perfscore{40}{83.3} & \perfscore{40}{76.7} & \perfscore{40}{46.7} & \perfscore{40}{58.0} & \perfscore{40}{30.8} \\
    \rowcolor{gray!15}
    +\ours{} & $y_r$ & \textbf{\perfscore{100}{63.1}} & \textbf{\perfscore{100}{49.4}} & \textbf{\perfscore{100}{32.1}} & \perfscore{80}{32.7} & \textbf{\perfscore{100}{11.2}} & \textbf{\perfscore{100}{86.7}} & \textbf{\perfscore{100}{80.0}} & \textbf{\perfscore{100}{56.7}} & \textbf{\perfscore{100}{64.0}} & \textbf{\perfscore{100}{33.8}} \\
    \midrule
    \multicolumn{12}{l}{\textit{Qwen3-8B}} \\
    \rowcolor{gray!15}
    Base & -- & \textcolor[HTML]{707070}{72.3} & \textcolor[HTML]{707070}{59.7} & \textcolor[HTML]{707070}{30.5} & \textcolor[HTML]{707070}{35.6} & \textcolor[HTML]{707070}{8.9} & \textcolor[HTML]{707070}{83.3} & \textcolor[HTML]{707070}{76.6} & \textcolor[HTML]{707070}{66.7} & \textcolor[HTML]{707070}{62.0} & \textcolor[HTML]{707070}{41.0} \\
    +OPSD & $y_o$ & \perfscore{40}{74.8} & \perfscore{40}{65.4} & \perfscore{40}{40.3} & \perfscore{40}{39.6} & \perfscore{40}{15.7} & \perfscore{70}{86.7} & \perfscore{40}{83.3} & \perfscore{81}{73.3} & \perfscore{40}{68.0} & \perfscore{40}{51.3} \\
    \rowcolor{gray!15}
    +OPSD & $y_t$ & \textbf{\perfscore{100}{77.9}} & \textbf{\perfscore{100}{69.2}} & \perfscore{69}{42.3} & \perfscore{51}{40.2} & \perfscore{59}{17.0} & \perfscore{40}{83.3} & \textbf{\perfscore{100}{86.7}} & \perfscore{40}{66.7} & \perfscore{70}{69.0} & \perfscore{40}{51.3} \\
    +OPSD & $y^\star$ & \perfscore{40}{74.8} & \perfscore{98}{69.1} & \perfscore{93}{44.0} & \perfscore{91}{42.3} & \textbf{\perfscore{100}{19.9}} & \textbf{\perfscore{100}{90.0}} & \textbf{\perfscore{100}{86.7}} & \perfscore{40}{66.7} & \textbf{\perfscore{100}{70.0}} & \perfscore{70}{56.4} \\
    \rowcolor{gray!15}
    +\ours{} & $y_r$ & \perfscore{73}{76.5} & \textbf{\perfscore{100}{69.2}} & \textbf{\perfscore{100}{44.5}} & \textbf{\perfscore{100}{42.8}} & \perfscore{63}{17.3} & \textbf{\perfscore{100}{90.0}} & \textbf{\perfscore{100}{86.7}} & \textbf{\perfscore{100}{76.3}} & \perfscore{40}{68.0} & \textbf{\perfscore{100}{61.5}} \\
    \bottomrule
  \end{tabular}%
  }
\end{table}

\paragraph{Trajectory Sources and the Support Constraint.}
To examine the role of the support constraint, we hold the sampling budget fixed and vary the trajectory used for the OPSD update: the raw student rollout $y_o$, a trajectory $y_t$ generated by Qwen3-14B, the reference solution $y^\star$, or the refined trajectory $y_r$. Here, Qwen3-14B supplies only $y_t$; the subsequent distillation update still uses the privileged self-teacher. Unlike student rollouts, $y_t$ and $y^\star$ are not constrained to the current policy's support and may therefore violate the constraint in \cref{eq:prefix-refinement-objective}.

\Cref{tab:trajectory-source-comparison} shows that these alternative sources help on individual benchmarks but do not match refinement overall. The refined trajectory $y_r$ is best or tied-best in $16$ of $20$ settings, including $9$ of $10$ Pass@16 settings. There are exceptions: on Qwen3-8B, $y_t$ achieves higher AIME24 Avg@16 than $y_r$ ($77.9\%$ vs. $76.5\%$), and $y^\star$ achieves higher AMOBench Avg@16 ($19.9\%$ vs. $17.3\%$). The overall pattern supports improving trajectory quality while retaining a connection to the student's own rollouts. This comparison is consistent with the support-constrained formulation, although it does not isolate support mismatch from other differences between trajectory sources.

\paragraph{Best-of-$N$ on the Full Training Corpus.}
Best-of-$N$ filtering offers a complementary way to pursue the objective while satisfying the support constraint: draw $N$ trajectories from the current student and retain verifier-passing outputs. Increasing $N$ raises the chance of obtaining a successful trajectory for each training problem. In \cref{fig:eq5-best-of-n-scaling}, we sample from the full Qwen3-4B-Instruct-2507 training corpus and allow sampling cost to grow with $N$. Mean Avg@16 increases from $35.66\%$ for Best-of-1 to $36.68\%$ for Best-of-8, compared with $36.34\%$ for vanilla OPSD. These gains support the usefulness of successful on-policy trajectories for downstream learning, consistent with \cref{eq:prefix-refinement-objective} as a surrogate for mitigating prefix failure.

The gains come at a substantial compute cost, as anticipated in \cref{sec:method}. Best-of-8 uses $42.9$ hours of total wall-clock time compared with $5.5$ hours for \ours{}, approximately $7.8\times$ as much, yet remains below \ours{} on four benchmarks and ties only on BeyondAIME. Its mean Avg@16 also falls short of the $37.70\%$ achieved by \ours{}. Thus, verifier-guided selection can improve the final policy, but refinement uses privileged guidance to obtain stronger supervision more efficiently than repeated unassisted sampling.

\begin{table}[t]
  \caption{Matched-budget OPSD with $N\in\{1,4,8\}$ sampled rollouts per problem in the privileged self-teacher regime (\%). OPSD ($N=1$) is the vanilla baseline; OPSD ($N=4$) and OPSD ($N=8$) retain only verifier-passing $y_o$ trajectories. \ours{} trains on $y_r$. Bold indicates the best or tied-best result in each model block.}
  \label{tab:matched-best-of-n}
  \centering
  \scriptsize
  \setlength{\tabcolsep}{1.8pt}
  \resizebox{\textwidth}{!}{%
  \begin{tabular}{lc*{10}{c}}
    \toprule
    & & \multicolumn{5}{c}{Avg@16} & \multicolumn{5}{c}{Pass@16} \\
    \cmidrule(lr){3-7}\cmidrule(lr){8-12}
    Method & Traj. & AIME24 & AIME25 & HMMT25 & BeyondAIME & AMOBench & AIME24 & AIME25 & HMMT25 & BeyondAIME & AMOBench \\
    \midrule
    \multicolumn{12}{l}{\textit{Qwen3-4B-Instruct-2507}} \\
    \rowcolor{gray!15}
    Base & -- & \textcolor[HTML]{707070}{59.1} & \textcolor[HTML]{707070}{42.9} & \textcolor[HTML]{707070}{28.3} & \textcolor[HTML]{707070}{28.2} & \textcolor[HTML]{707070}{8.1} & \textcolor[HTML]{707070}{73.3} & \textcolor[HTML]{707070}{66.7} & \textcolor[HTML]{707070}{43.3} & \textcolor[HTML]{707070}{54.0} & \textcolor[HTML]{707070}{23.1} \\
    +OPSD ($N=1$) & $y_o$ & \perfscore{40}{60.3} & \perfscore{88}{48.8} & \perfscore{69}{30.8} & \perfscore{60}{32.3} & \perfscore{54}{9.5} & \perfscore{40}{86.7} & \perfscore{40}{76.7} & \perfscore{70}{53.3} & \perfscore{40}{58.0} & \perfscore{98}{33.3} \\
    \rowcolor{gray!15}
    +OPSD ($N=4$) & $y_o$ & \perfscore{66}{61.5} & \perfscore{52}{47.1} & \perfscore{40}{29.6} & \textbf{\perfscore{100}{32.7}} & \perfscore{40}{9.0} & \textbf{\perfscore{100}{90.0}} & \perfscore{40}{76.7} & \perfscore{40}{50.0} & \perfscore{50}{59.0} & \perfscore{40}{20.5} \\
    +OPSD ($N=8$) & $y_o$ & \perfscore{46}{60.6} & \perfscore{40}{46.5} & \perfscore{54}{30.2} & \perfscore{40}{32.1} & \perfscore{84}{10.6} & \perfscore{40}{86.7} & \perfscore{40}{76.7} & \textbf{\perfscore{100}{56.7}} & \perfscore{70}{61.0} & \perfscore{98}{33.3} \\
    \rowcolor{gray!15}
    +\ours{} & $y_r$ & \textbf{\perfscore{100}{63.1}} & \textbf{\perfscore{100}{49.4}} & \textbf{\perfscore{100}{32.1}} & \textbf{\perfscore{100}{32.7}} & \textbf{\perfscore{100}{11.2}} & \perfscore{40}{86.7} & \textbf{\perfscore{100}{80.0}} & \textbf{\perfscore{100}{56.7}} & \textbf{\perfscore{100}{64.0}} & \textbf{\perfscore{100}{33.8}} \\
    \midrule
    \multicolumn{12}{l}{\textit{Qwen3-8B}} \\
    \rowcolor{gray!15}
    Base & -- & \textcolor[HTML]{707070}{72.3} & \textcolor[HTML]{707070}{59.7} & \textcolor[HTML]{707070}{30.5} & \textcolor[HTML]{707070}{35.6} & \textcolor[HTML]{707070}{8.9} & \textcolor[HTML]{707070}{83.3} & \textcolor[HTML]{707070}{76.6} & \textcolor[HTML]{707070}{66.7} & \textcolor[HTML]{707070}{62.0} & \textcolor[HTML]{707070}{41.0} \\
    +OPSD ($N=1$) & $y_o$ & \perfscore{59}{74.8} & \perfscore{40}{65.4} & \perfscore{40}{40.3} & \perfscore{40}{39.6} & \perfscore{40}{15.7} & \perfscore{70}{86.7} & \perfscore{40}{83.3} & \perfscore{81}{73.3} & \textbf{\perfscore{100}{68.0}} & \perfscore{66}{51.3} \\
    \rowcolor{gray!15}
    +OPSD ($N=4$) & $y_o$ & \perfscore{78}{75.6} & \perfscore{86}{68.3} & \perfscore{96}{44.2} & \perfscore{86}{42.3} & \perfscore{96}{17.2} & \perfscore{40}{83.3} & \textbf{\perfscore{100}{86.7}} & \perfscore{40}{66.7} & \perfscore{40}{67.0} & \perfscore{40}{43.6} \\
    +OPSD ($N=8$) & $y_o$ & \perfscore{40}{74.0} & \perfscore{94}{68.8} & \perfscore{93}{44.0} & \textbf{\perfscore{100}{43.1}} & \perfscore{89}{17.0} & \perfscore{70}{86.7} & \textbf{\perfscore{100}{86.7}} & \perfscore{40}{66.7} & \textbf{\perfscore{100}{68.0}} & \perfscore{66}{51.3} \\
    \rowcolor{gray!15}
    +\ours{} & $y_r$ & \textbf{\perfscore{100}{76.5}} & \textbf{\perfscore{100}{69.2}} & \textbf{\perfscore{100}{44.5}} & \perfscore{95}{42.8} & \textbf{\perfscore{100}{17.3}} & \textbf{\perfscore{100}{90.0}} & \textbf{\perfscore{100}{86.7}} & \textbf{\perfscore{100}{76.3}} & \textbf{\perfscore{100}{68.0}} & \textbf{\perfscore{100}{61.5}} \\
    \bottomrule
  \end{tabular}%
  }
\end{table}

\paragraph{OPSD under a Matched Sampling Budget.}
We also fix the sampling budget to compare OPSD with $N\in\{1,4,8\}$ rollouts per problem (\cref{tab:matched-best-of-n}). OPSD with $N=1$ is the vanilla baseline, while the $N=4$ and $N=8$ variants retain only verifier-passing student trajectories. Allocating more draws to each problem therefore trades coverage of distinct training problems for a higher chance of finding a successful trajectory. OPSD with $N=8$ matches or exceeds the $N=1$ baseline in $15$ of $20$ settings, but remains below \ours{} overall. Across all compared methods, \ours{} is best or tied-best in $18$ of $20$ settings; the exceptions are Qwen3-4B AIME24 Pass@16 and Qwen3-8B BeyondAIME Avg@16.

The two sampling regimes expose the same practical limitation: Best-of-$N$ must spend additional draws to discover successful trajectories and provides no supervision for problems where all sampled candidates fail. \ours{} instead uses a failed rollout as context for privileged refinement, allowing those problems to contribute supervision through $y_r$. The results support optimizing the trajectory-construction objective while favoring refinement as the more effective use of limited sampling resources.

\subsection{Trajectory Analysis}
\label[appendix]{subsec:trajectory-analysis}

We analyze training trajectories and correct evaluation rollouts to characterize how trajectory refinement changes supervision and solution length, complementing the coverage analysis in \cref{subsec:exploration}.

\begin{figure}[!h]
\centering
\includegraphics[width=0.82\linewidth,trim=0bp 20bp 166.2bp 16bp,clip]{figs/trajectory_rollout_three_panel.pdf}
\caption{Trajectory evidence for the construction objective. \textbf{Left}: privileged self-teacher training-corpus trajectory length on Qwen3-8B, with the orange line showing verifier accuracy. \textbf{Right}: separate-teacher AMOBench correct-rollout length distribution for Qwen3-4B-Instruct-2507.}
\label{fig:trajectory-rollout-analysis}
\end{figure}

\paragraph{Training Trajectory Analysis ($y_o \text{ vs. } y_r$).}
\label[appendix]{exp:trajectory}
In the privileged self-teacher regime, the left panel of \cref{fig:trajectory-rollout-analysis} contrasts $y^\star$, the raw rollout $y_o$, and the refined trajectory $y_r$ over the Qwen3-8B training corpus. The constructed proposal increases the verifier-pass rate from $65.8\%$ on $y_o$ to $81.4\%$ on $y_r$, while compressing the median length by roughly $9\times$, from $7.7$K to $0.88$K, toward the reference scale of roughly $0.49$K. The same compression appears on the correct subset of $y_o$, where $y_r$ exposes the student to alternative short derivations rather than only repairing failures. The shorter $y_r$ also reduces Qwen3-8B training wall-clock by roughly $60\%$, partially offsetting the additional refinement pass (\cref{app:hardware}).

\paragraph{Rollout Trajectory Analysis.}
\label[appendix]{exp:rollout-analysis}
We further analyze the AMOBench evaluation rollouts for Qwen3-4B-Instruct-2507 under the separate-teacher regime, using the same $K{=}128$ samples per question as in \cref{subsec:exploration}. The right panel of \cref{fig:trajectory-rollout-analysis} shows that vanilla OPD and \ours{} produce highly similar correct-rollout length distributions, while \ours{} is slightly shorter on average ($18.9$K to $18.5$K characters) with a modest density shift toward the $18$K to $20$K range. The broader coverage in \cref{fig:exploration-passk} therefore accompanies only a small change in the lengths of successful solutions.

\subsection{Test Rollout Analysis}
\label[appendix]{app:test-rollout}

\paragraph{Setup.}
We analyze AMOBench test rollouts from the Qwen3-8B checkpoints evaluated under OPSD in \cref{tab:opsd-avg16,tab:opsd-pass16}. We compare the base model, vanilla forward-KL OPSD, and \ours{}, using $K{=}16$ completions per question and the generation and verification protocol in \cref{exp:setup}.

\paragraph{Base-difficulty buckets.}
Let $b_q \in \{0,\ldots,16\}$ denote the number of correct base-model completions for question $q$. We partition the questions into three buckets: B$0$ ($n=23$, $b_q=0$), B$1$ to $8$ ($n=12$, $1\leq b_q\leq 8$), and B$9$ to $16$ ($n=4$, $9\leq b_q\leq 16$). The buckets are determined by the base model and held fixed when evaluating both trained methods.

\begin{figure}[!htbp]
\centering
\includegraphics[width=0.5\linewidth]{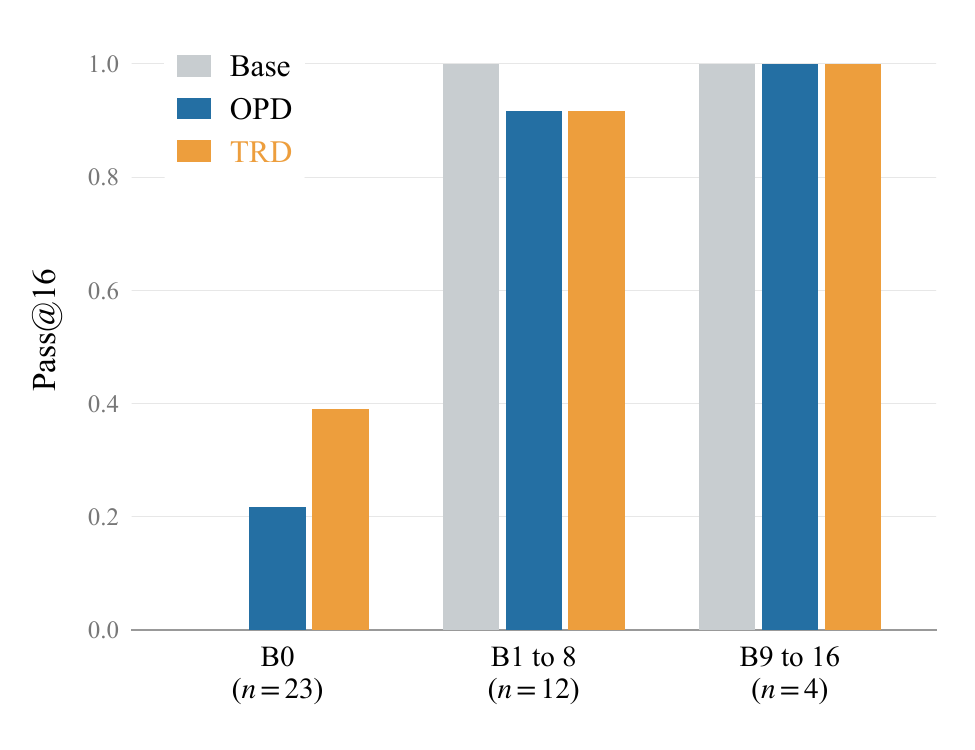}
\caption{AMOBench Pass@16 by base-model difficulty for Qwen3-8B under OPSD, with $K{=}16$ completions per question. The buckets contain questions with $0$, $1$ to $8$, or $9$ to $16$ correct base-model completions, respectively. B$0$ contains the $23$ questions left unsolved by the base model in all $16$ attempts. OPD denotes the vanilla forward-KL baseline.}
\label{fig:test-rollout-analysis}
\end{figure}

\paragraph{Coverage on Questions Unsolved by the Base Model.}
\Cref{fig:test-rollout-analysis} shows that \ours{} solves $9$ of the $23$ questions in B$0$, achieving Pass@16 $0.39$, compared with $0.22$ for vanilla forward-KL OPSD, the strongest dense-KL baseline on AMOBench Pass@16 in \cref{tab:opsd-pass16}. This nearly doubles the baseline's coverage on the subset left unsolved by the base model under the same sampling budget.

\paragraph{Coverage on the Remaining Questions.}
In B$1$ to $8$, both vanilla forward-KL OPSD and \ours{} achieve Pass@16 $0.92$, compared with $1.00$ for the base model. All three methods achieve $1.00$ in B$9$ to $16$. The coverage advantage of \ours{} over vanilla forward-KL OPSD is therefore concentrated in B$0$.

\subsection{Comparing Separate and Privileged Self-Teachers}
\label[appendix]{subsec:opd-opsd-trd}

\begin{table}[!h]
  \caption{TRD comparison between the separate Qwen3-8B teacher and the privileged self-teacher on Qwen3-4B-Instruct-2507 math benchmarks. Bold marks the better value between the two teacher regimes. Teacher scores are gray.}
  \label{tab:opd-opsd-teacher-4b}
  \centering
  \scriptsize
  \setlength{\tabcolsep}{2.5pt}
  \resizebox{\textwidth}{!}{%
  \begin{tabular}{lcccccccccc}
    \toprule
    & \multicolumn{5}{c}{Avg@16} & \multicolumn{5}{c}{Pass@16} \\
    \cmidrule(lr){2-6}\cmidrule(lr){7-11}
    Setting & AIME24 & AIME25 & HMMT25 & BeyondAIME & AMOBench & AIME24 & AIME25 & HMMT25 & BeyondAIME & AMOBench \\
    \midrule
    \rowcolor{gray!15}
    Teacher & \textcolor[HTML]{707070}{72.3} & \textcolor[HTML]{707070}{59.7} & \textcolor[HTML]{707070}{30.5} & \textcolor[HTML]{707070}{35.6} & \textcolor[HTML]{707070}{8.9} & \textcolor[HTML]{707070}{83.3} & \textcolor[HTML]{707070}{76.6} & \textcolor[HTML]{707070}{66.7} & \textcolor[HTML]{707070}{62.0} & \textcolor[HTML]{707070}{41.0} \\
    \midrule
    Separate teacher & \textbf{\perfscore{100}{65.4}} & \perfscore{40}{47.9} & \textbf{\perfscore{100}{33.2}} & \perfscore{40}{32.6} & \perfscore{40}{10.3} & \perfscore{40}{83.3} & \perfscore{40}{76.7} & \perfscore{40}{50.0} & \perfscore{40}{62.0} & \textbf{\perfscore{100}{35.9}} \\
    \rowcolor{gray!15}
    Privileged self-teacher & \perfscore{40}{63.1} & \textbf{\perfscore{100}{49.4}} & \perfscore{40}{32.1} & \textbf{\perfscore{100}{32.7}} & \textbf{\perfscore{100}{11.2}} & \textbf{\perfscore{100}{86.7}} & \textbf{\perfscore{100}{80.0}} & \textbf{\perfscore{100}{56.7}} & \textbf{\perfscore{100}{64.0}} & \perfscore{40}{33.8} \\
    \bottomrule
  \end{tabular}%
  }
\end{table}

\Cref{tab:opd-opsd-teacher-4b} compares Avg@16 and Pass@16 between the two teacher regimes for \ours{} at the same Qwen3-4B-Instruct scale. Avg@16 is mixed: the separate teacher is stronger on AIME24 and HMMT25, while the privileged self-teacher is stronger on AIME25, BeyondAIME, and AMOBench. Pass@16 is generally stronger with the privileged self-teacher, which wins on four of five competition-math benchmarks.

The gain suggests that, for optimizing \cref{eq:prefix-refinement-objective}, privileged information can be more effective than teacher scaling: some questions may remain beyond the separate teacher's ability to refine, while the reference directly supplies the correct solution structure. Using a stronger separate teacher may reduce such refinement failures, but increases computational cost. Besides, because the privileged self-teacher refines with the student backbone conditioned on the reference, it potentially stays closer to the student's support and avoids mismatch between models \citep{fu2026revisiting,li2026rethinking}.

\section{Experiment Details}
\label[appendix]{app:experiment-details}

This appendix follows the experimental workflow. We first describe training data and trajectory construction (\cref{app:data}), followed by the OPD/OPSD and GRPO training configurations (\cref{app:distillation-hparams,app:grpo}). We then specify compute accounting, training diagnostics, and evaluation (\cref{app:hardware,app:training-metrics,app:eval}). Complete initial-response and refinement prompts are collected in \cref{app:refinement-prompt}.

\subsection{Training Data and Trajectory Construction}
\label[appendix]{app:data}

Training uses DeepScaleR for math~\citep{deepscaler2025} and TACO for code~\citep{li2023taco}. For OPD/OPSD, Stage 1 samples a raw student rollout $y_o$ conditioned on the problem; \ours{} adds Stage 2, in which the teacher refines $y_o$ into $y_r$. We train on the selected trajectories for one epoch and merge the LoRA adapters before evaluation. \Cref{tab:training-stages} lists the generation settings, and \cref{app:refinement-prompt} provides the complete prompts.

\begin{table}[h]
\caption{Training generation configuration. Stage 1 constructs $y_o$ for the OPD/OPSD variants; Stage 2 constructs $y_r$ only for \ours{}.}
\label{tab:training-stages}
\centering
\scriptsize
\setlength{\tabcolsep}{5pt}
\begin{tabular}{>{\raggedright\arraybackslash}p{0.22\linewidth}>{\raggedright\arraybackslash}p{0.22\linewidth}>{\raggedright\arraybackslash}p{0.22\linewidth}>{\raggedright\arraybackslash}p{0.22\linewidth}}
\toprule
Setting & Stage 1: $y_o$ & OPD Stage 2: $y_r$ & OPSD Stage 2: $y_r$ \\
\midrule
Samples per problem & $1$ & $1$ & $1$ \\
Temperature & $0.6$ & $0.6$ & $0.6$ \\
Top-$p$ & $0.95$ & $0.95$ & $0.95$ \\
Top-$k$ & $20$ & $20$ & $20$ \\
Prompt budget & Math: $4{,}096$; code: $2{,}048$ tokens & $18{,}432$ tokens & $22{,}528$ tokens \\
Response budget & $16{,}384$ tokens & $16{,}384$ tokens & $16{,}384$ tokens \\
Maximum model length & Math: $20{,}480$; code: $18{,}432$ tokens & $34{,}816$ tokens & $38{,}912$ tokens \\
\bottomrule
\end{tabular}
\end{table}

\subsection{OPD/OPSD Hyperparameters}
\label[appendix]{app:distillation-hparams}

\paragraph{Models and teacher conditioning.}
OPD uses a frozen separate Qwen3-8B teacher and Qwen3-1.7B or Qwen3-4B-Instruct-2507 students~\citep{yang2025qwen3}. The teacher branch is never updated and its logits are detached before KL computation. OPSD uses Qwen3-4B-Instruct-2507 and Qwen3-8B as shared teacher/student backbones: the same base model produces privileged teacher logits under reference-solution conditioning, while LoRA adapters update only the student branch.

OPD and OPSD share the math/code corpora, prompt adapters, Stage-1 rollout sampler, KL implementation, optimizer, LoRA configuration, FSDP2 execution path, and evaluation sampling parameters. They differ in teacher identity, the teacher's access to $y^\star$, and the longer OPSD prompt budgets needed to include $y^\star$.

\paragraph{Shared optimization.}
All OPD/OPSD variants and \ours{} use AdamW, bfloat16, gradient checkpointing, and one training epoch, with the shared settings in \cref{tab:common-hparams}. On one $8$-GPU node, the per-GPU batch is $1$ and gradient accumulation is $16$, giving effective batch $128$.

We apply LoRA of rank $r{=}64$, scaling $\alpha{=}128$, dropout $0.05$, on all attention and MLP linear layers (\texttt{q\_proj}, \texttt{k\_proj}, \texttt{v\_proj}, \texttt{o\_proj}, \texttt{gate\_proj}, \texttt{up\_proj}, \texttt{down\_proj}). After training we merge the adapters into the base weights before evaluation.

\begin{table}[h]
\caption{Common optimization and generation hyperparameters used by the direct OPD/OPSD recipes.}
\label{tab:common-hparams}
\centering
\scriptsize
\setlength{\tabcolsep}{6pt}
\begin{tabular}{ll}
\toprule
Setting & Value \\
\midrule
Optimizer            & AdamW ($\beta_1{=}0.9$, $\beta_2{=}0.999$, $\epsilon{=}10^{-8}$) \\
Peak learning rate   & $5\!\times\!10^{-6}$ \\
Precision            & bfloat16 \\
Gradient checkpointing & enabled \\
Per-GPU train batch  & $1$ \\
Gradient accumulation & $16$ \\
Gradient clipping    & $1.0$ \\
LR schedule          & linear warmup, cosine decay to $0.1\!\times$ peak LR \\
Warmup ratio         & $0.1$ \\
Weight decay         & $0.005$ \\
Epochs per update    & $1$ \\
Full-vocab KL chunk size & $512$ tokens \\
LoRA save/merge      & save adapter checkpoints and merge before evaluation \\
Sequence packing     & remove-padding via verl FSDP2 \\
Sequence parallel    & ulysses, size $1$ \\
Rollout generation   & temperature $0.6$, top-$p{=}0.95$, top-$k{=}20$, max sequences $64$ \\
\bottomrule
\end{tabular}
\end{table}

\paragraph{Distillation losses and method-specific settings.}
We use KL temperature $T{=}1.0$ and the Qwen3 vocabulary ($|\mathcal{V}|\!\approx\!152$K), with teacher top-$32$ truncation for the Top-$K$ variant. \Cref{tab:method-hparams} summarizes the trajectory, loss, teacher prompt, and context-length settings for each method.

Full-vocabulary variants compute the per-token KL at temperature $T$ as
\begin{equation}
D^{(T)}\big(p\,\|\,q\big) \;=\; \sum_{v\in\mathcal{V}} p_T(v)\,\big(\log p_T(v) - \log q_T(v)\big),
\end{equation}
with $p_T(v) \propto \exp(\ell_v / T)$. The clipping baseline applies a per-token cap $\min(D_{\mathrm{KL}, t}, c)$ before averaging over the response mask, with $c{=}0.1$ in both OPD and OPSD. Top-$K$ replaces $\mathcal{V}$ by the teacher's top-$32$ support $\mathcal{S}_t$ and renormalizes both $p_T$ and $q_T$ to sum to $1$ on $\mathcal{S}_t$ before evaluating $D$.

\begin{table}[h]
\caption{Method-specific OPD/OPSD settings. Clipped forward KL uses $c{=}0.1$ in both regimes. Max-length columns are training-time model lengths for teacher-forced KL.}
\label{tab:method-hparams}
\centering
\scriptsize
\setlength{\tabcolsep}{4pt}
\begin{tabular}{lcccccc}
\toprule
Method & Traj. & KL dir. & Clip $c$ & Top-$K$ & Teacher prompt & OPD / OPSD max len. \\
\midrule
Forward KL                  & $y_o$ & forward & $0$ & -- & vanilla & $18{,}432$ / $22{,}528$ \\
Forward KL w/ Clip          & $y_o$ & forward & $0.1$ & -- & vanilla & $18{,}432$ / $22{,}528$ \\
Reverse KL                  & $y_o$ & reverse & $0$ & -- & vanilla & $18{,}432$ / $22{,}528$ \\
Reverse KL w/ Top-$K$       & $y_o$ & reverse & $0$ & $32$ & vanilla & $18{,}432$ / $22{,}528$ \\
\ours{} (ours)              & $y_r$ & forward & $0$ & -- & refine & $34{,}816$ / $38{,}912$ \\
\bottomrule
\end{tabular}
\end{table}

\subsection{GRPO Algorithm and Hyperparameters}
\label[appendix]{app:grpo}

We implement GRPO~\citep{shao2024deepseekmath} in verl using the task-specific configurations.\footnote{The configurations are available in \href{https://github.com/louieworth/verl/tree/2c903ed478ebaad63a13034b3b19c2b73496e0ad/recipe/opd/run/grpo}{\texttt{recipe/opd/run/grpo}}.} For each problem, we sample $G{=}8$ responses from the current policy and compute a scalar reward $r_i$ for each response. The group-relative advantage is $A_i=(r_i-\bar r)/(s_r+10^{-6})$, where $\bar r$ and $s_r$ are the within-group reward mean and sample standard deviation. This advantage is shared across the response tokens. The policy uses a clipped token-level objective with token-mean aggregation and a KL penalty to the frozen initial policy, without a learned critic. The KL penalty is applied in the actor loss; entropy regularization is disabled. \Cref{tab:grpo-hparams} specifies the training configuration.

\paragraph{Rewards.}
For math, the reward is $0.67$ for a correct extracted answer plus $0.33$ for satisfying the prescribed \texttt{<think>} and \texttt{<answer>} format with a boxed answer. For code, the reward is $1$ when all TACO tests pass; otherwise, it is the fraction passed among the first $10$ tests (or all tests when fewer are available). Reference answers and tests are used for reward computation, while the policy is conditioned on the problem.

\begin{table}[h]
\caption{GRPO training hyperparameters. }
\label{tab:grpo-hparams}
\centering
\begin{minipage}{0.85\textwidth}
\centering
\scriptsize
\setlength{\tabcolsep}{5pt}
\begin{tabular}{>{\raggedright\arraybackslash}p{\dimexpr0.46\linewidth-2\tabcolsep\relax}>{\raggedright\arraybackslash}p{\dimexpr0.54\linewidth-2\tabcolsep\relax}}
\toprule
Setting & Value \\
\midrule
Parameter updates & Full model, bfloat16, FSDP2 \\
Optimizer & AdamW, $\beta_1{=}0.9$, $\beta_2{=}0.999$ \\
Learning rate & $10^{-6}$, constant, no warmup \\
Weight decay / gradient clipping & $0.01$ / $1.0$ \\
Rollouts per problem & $8$ \\
Problems per rollout batch & $64$ \\
Responses per optimizer minibatch & $256$ \\
Optimization epochs per rollout batch & $1$ \\
Policy clipping & Lower/upper widths $0.2$ \\
Reference-policy KL & Coefficient $0.001$ \\
Training sampler & Temperature $1.0$, top-$p{=}1.0$, top-$k{=}-1$ \\
Maximum prompt length & Math: $2{,}048$; code: $4{,}096$ tokens \\
Maximum response length & $16{,}384$ tokens \\
Dynamic microbatch token budget per GPU & $32{,}768$; $18{,}432$ for Qwen3-8B math \\
\bottomrule
\end{tabular}
\end{minipage}
\end{table}

The rollout-and-training budget follows the GPU-time matching protocol in \cref{app:hardware}.

\subsection{Hardware and Compute}
\label[appendix]{app:hardware}

All runs use a single node of $8\times$ H100 80GB GPUs with FSDP2 sharding via verl~\citep{sheng2024verl}.

\paragraph{GPU-time matching.}
We match GRPO and \ours{} by total GPU time covering all rollout generation and training, including both the raw-rollout and refinement stages for \ours{}. With the same 8 H100 GPUs, this is equivalent to matching the elapsed rollout-and-training time. The GRPO timer starts before the first rollout, and training stops after the current update completes once the time budget is reached, so matching is subject to the granularity of one update. Initialization, checkpoint saving, and final evaluation are outside this budget.

\paragraph{Measured wall-clock.}
\cref{tab:wallclock} reports the training-pipeline wall-clock by model, setting, and method. The main trade-off is that \ours{} adds an extra sampling pass to construct $y_r$, but this overhead is partially offset by faster KL training because the refined trajectories are much shorter than $y_o$ (see \cref{subsec:trajectory-analysis}). This offset becomes more pronounced as the backbone grows: on Qwen3-8B, \ours{} and Vanilla OPSD have nearly matched total wall-clock ($9{:}20$ vs.\ $9{:}40$).

\begin{table}[h]
\caption{Approximate OPD/OPSD training-pipeline wall-clock on a single $8\times$ H100 80GB node. The $y_o$ and $y_r$ columns report rollout generation time; $y_r$ is used only by \ours{}. Training covers the KL update only. Times are rounded to $10$-minute bins.}
\label{tab:wallclock}
\centering
\scriptsize
\setlength{\tabcolsep}{4pt}
\begin{tabular}{>{\raggedright\arraybackslash}p{0.25\linewidth}llcccc}
\toprule
Model & Setting & Method & $y_o$ rollout & $y_r$ rollout & Training & Total \\
\midrule
\multirow{2}{=}{Qwen3-1.7B}
  & \multirow{2}{*}{OPD} & Vanilla OPD & $3{:}30$ & -- & $1{:}10$ & $4{:}40$ \\
  & & \ours{} & $3{:}30$ & $4{:}00$ & $0{:}40$ & $8{:}10$ \\
\midrule
\multirow{4}{=}{Qwen3-4B-Instruct-2507}
  & \multirow{2}{*}{OPD} & Vanilla OPD & $4{:}00$ & -- & $1{:}20$ & $5{:}20$ \\
  & & \ours{} & $4{:}00$ & $4{:}00$ & $1{:}00$ & $9{:}00$ \\
\cmidrule(lr){2-7}
  & \multirow{2}{*}{OPSD} & Vanilla OPSD & $2{:}10$ & -- & $2{:}10$ & $4{:}20$ \\
  & & \ours{} & $2{:}10$ & $2{:}00$ & $1{:}20$ & $5{:}30$ \\
\midrule
\multirow{2}{=}{Qwen3-8B}
  & \multirow{2}{*}{OPSD} & Vanilla OPSD & $4{:}20$ & -- & $5{:}20$ & $9{:}40$ \\
  & & \ours{} & $4{:}20$ & $2{:}50$ & $2{:}10$ & $9{:}20$ \\
\bottomrule
\end{tabular}
\end{table}

\subsection{Training Diagnostics}
\label[appendix]{app:training-metrics}

The diagnostic curves in \cref{fig:prefix-failure-evidence}, discussed in \cref{subsec:exp_prefix}, use the privileged self-teacher setting to control for differences in teacher and student backbone models. The same trainer can log these metrics for OPD, but the reported OPD direct-variant runs keep them off unless explicitly enabled. Let $z_i$ denote the response scored in a run: $z_i=y_{o,i}$ for standard OPSD and $z_i=y_{r,i}$ for \ours{}. We use response-token masks $m_{i,t}\in\{0,1\}$ and aggregate within each logging step. KL and epistemic-token mass are token-weighted means; perplexity is the exponential of a token-weighted mean negative log-likelihood. Prompt conditioning follows \cref{sec:method} and is suppressed in the probabilities below: the student receives the problem, while the teacher receives privileged reference information and, for \ours{}, the original rollout.

\paragraph{Per-token KL by rollout outcome.}
Each prompt $x_i$ carries a fixed binary stage-1 outcome label $b_i$, set to $1$ (\emph{correct}) if the base model's original rollout $y_{o,i}$ passes the verifier and to $0$ (\emph{incorrect}) otherwise. The paired refined response $y_{r,i}$ inherits this label, so the bucket names do not describe the refined response's own correctness. The per-token KL between teacher and student is averaged separately within each bucket,
\begin{equation}
\begin{aligned}
D^{\text{correct}} &\;=\; \frac{\sum_{i:\,b_i=1}\sum_t D_{\mathrm{KL},i,t}\,m_{i,t}}{\sum_{i:\,b_i=1}\sum_t m_{i,t}},
\\
D^{\text{incorrect}} &\;=\; \frac{\sum_{i:\,b_i=0}\sum_t D_{\mathrm{KL},i,t}\,m_{i,t}}{\sum_{i:\,b_i=0}\sum_t m_{i,t}},
\end{aligned}
\end{equation}
where $D_{\mathrm{KL},i,t}$ uses the KL direction configured for that run and is evaluated at prefix $z_{i,<t}$. These are means per unmasked token, rather than trajectory sums. Normalizing by token count removes the direct scaling with the number of evaluated tokens, but does not equate the underlying prefix distributions or the forward- and reverse-KL objectives.

\paragraph{Teacher-student perplexity gap.}
Both perplexities use teacher-forced decoding on the same response tokens and mask within each run,
\begin{equation}
\begin{aligned}
\mathrm{PPL}_S &\;=\; \exp\!\Bigg(\frac{\sum_{i,t} -\log \pi_S(z_{i,t}\mid z_{i,<t})\, m_{i,t}}{\sum_{i,t} m_{i,t}}\Bigg),
\\
\mathrm{PPL}_T &\;=\; \exp\!\Bigg(\frac{\sum_{i,t} -\log \pi_T(z_{i,t}\mid z_{i,<t})\, m_{i,t}}{\sum_{i,t} m_{i,t}}\Bigg).
\end{aligned}
\end{equation}
We report $\mathrm{PPL}_S-\mathrm{PPL}_T$ using $T{=}1$ log-probabilities. This convention gives a consistent metric definition across runs, but runs scored on $y_o$ and $y_r$ evaluate different responses. Their gaps therefore describe likelihood differences on the respective response distributions, rather than a controlled comparison of supervision quality on a shared evaluation corpus.

\paragraph{Epistemic-token mass.}
Let $\mathcal{E}\subset\mathcal{V}$ be the set of \emph{epistemic onset} tokens. We construct $\mathcal{E}$ from $16$ phrases,
\begin{center}
\texttt{Wait}, \texttt{Actually}, \texttt{However}, \texttt{Alternatively}, \texttt{Oops}, \texttt{Wrong}, \texttt{Error}, \texttt{Incorrect}, \texttt{Correction}, \texttt{Sorry}, \texttt{Hmm}, \texttt{Oh}, \texttt{Hold}, \texttt{Pause}, \texttt{Uh}, \texttt{Um},
\end{center}
by tokenising each phrase under the student tokenizer in two variants (bare string and leading-space) and collecting the first sub-word id, then deduplicating. Thus, $16$ is the number of source phrases, not necessarily the number of distinct token ids. At each response position $(i,t)$ we measure how much teacher mass falls on $\mathcal{E}$ before any KL temperature scaling,
\begin{equation}
\mathrm{mass}_{i,t} \;=\; \sum_{v\in\mathcal{E}} \pi_T(v\mid z_{i,<t})
\;=\; \sum_{v\in\mathcal{E}} \mathrm{softmax}\!\big(\boldsymbol{\ell}^{T}_{i,t}\big)_v,
\end{equation}
and report the token-weighted mean $\big(\sum_{i,t} \mathrm{mass}_{i,t}\,m_{i,t}\big)\big/\big(\sum_{i,t} m_{i,t}\big)$. The metric is computed only on the full-vocabulary path (i.e., when teacher logits are materialised) and is independent of the KL-loss temperature.

\subsection{Evaluation Settings}
\label[appendix]{app:eval}

\paragraph{Benchmarks.}
The math suite contains AIME24/25~\citep{maa2024aime,maa2025aime}, HMMT25~\citep{hmmt2025}, BeyondAIME~\citep{seed2025beyondaime}, and the $39$ parser-graded AMOBench problems~\citep{amobench2025}. The code suite contains HumanEval+, MBPP+~\citep{evalplus}, and LiveCodeBench v6~\citep{jain2024livecodebench}.

\paragraph{Sampling and verification.}
\cref{tab:eval-config} summarizes the evaluation configuration. We use $K{=}16$ completions per problem; temperature and response length follow the Qwen3 evaluation setup~\citep{yang2025qwen3}. OPD/OPSD code evaluation uses up to $128$ concurrent sequences.

\begin{table}[h]
\caption{Evaluation configuration. The $K{=}16$ sampling budget is our reporting protocol for Avg@16 and Pass@16.}
\label{tab:eval-config}
\centering
\scriptsize
\setlength{\tabcolsep}{5pt}
\begin{tabular}{>{\raggedright\arraybackslash}p{0.22\linewidth}>{\raggedright\arraybackslash}p{0.34\linewidth}>{\raggedright\arraybackslash}p{0.34\linewidth}}
\toprule
Setting & Math evaluation & Code evaluation \\
\midrule
Samples per problem & $K{=}16$ & $K{=}16$ \\
Temperature & $0.6$ & $0.6$ \\
Top-$p$ & $0.95$ & $0.95$ \\
Top-$k$ & $20$ & $20$ \\
Prompt budget & $4{,}096$ tokens & $2{,}048$ tokens \\
Response budget & $38{,}912$ tokens & $16{,}384$ tokens \\
Maximum model length & $43{,}008$ tokens & $18{,}432$ tokens \\
Verifier & rule-based boxed-answer verifier & benchmark unit-test executor \\
\bottomrule
\end{tabular}
\end{table}

Math completions are scored with answer extraction from the final \texttt{\textbackslash boxed\{\ldots\}} block. HumanEval+ and MBPP+ are evaluated through EvalPlus; LiveCodeBench uses the \texttt{lcb\_runner} code-generation scenario with release version $6$.

\paragraph{Metrics.}
For each test question $x_q$ we draw $K{=}16$ independent completions $y_q^{(i)}\sim\pi_\theta(\cdot\mid x_q)$ and score them with a verifier $\operatorname{Verify}(\cdot,\cdot)$: the boxed-answer verifier for math and unit tests for code. Over $N$ test questions,
\begin{equation*}
\operatorname{Avg}@K \;=\; \frac{1}{N}\sum_{q=1}^{N}\frac{1}{K}\sum_{i=1}^{K}\operatorname{Verify}\!\big(y_q^{(i)},\,y_q^\star\big),
\quad
\operatorname{Pass}@K \;=\; \frac{1}{N}\sum_{q=1}^{N}\max_{1\le i\le K}\operatorname{Verify}\!\big(y_q^{(i)},\,y_q^\star\big).
\end{equation*}
Avg@16 tracks average sample quality; Pass@16 tracks whether at least one of the $16$ samples solves the problem.

\paragraph{Evaluation cost.}
\Cref{tab:eval-wallclock} reports evaluation time separately from rollout generation and training.

\begin{table}[h]
\caption{Approximate evaluation wall-clock on a single $8\times$ H100 80GB node. Times are rounded to $30$-minute bins and reported by model only.}
\label{tab:eval-wallclock}
\centering
\scriptsize
\setlength{\tabcolsep}{8pt}
\begin{tabular}{lcc}
\toprule
Model & Math suite & Code suite \\
\midrule
Qwen3-1.7B & $2{:}00$ & $6{:}30$ \\
Qwen3-4B-Instruct-2507 & $2{:}00$ & $3{:}00$ \\
Qwen3-8B & $4{:}00$ & -- \\
\bottomrule
\end{tabular}
\end{table}

\subsection{Initial and Refinement Prompts}
\label[appendix]{app:refinement-prompt}

Stage-1 initial-response prompts are task-specific but shared by OPD and OPSD. They produce the raw rollout $y_o$ used by the dense-KL baselines and by the Stage-2 refinement prompts. The refinement prompt is also task-specific and differs between OPD and OPSD only in whether the reference solution $y^\star$ is shown. OPD refinement uses the separate teacher and hides $y^\star$; OPSD refinement uses the shared model under privileged conditioning and includes $y^\star$.

\newcommand{\promptbox}[1]{%
\begin{center}
\setlength{\fboxsep}{7pt}%
\fcolorbox{coreblue}{corecolor}{%
\begin{minipage}{0.88\linewidth}
\vspace{0.6em}
\small
\setlength{\parindent}{0pt}
\setlength{\parskip}{0.35em}
#1
\vspace{0.45em}
\end{minipage}%
}%
\end{center}
}
\newcommand{\promptslot}[1]{\textcolor{coreblue}{\{\textsc{#1}\}}}
\newcommand{\promptfence}[1]{\texttt{\textasciigrave\textasciigrave\textasciigrave#1}}

\paragraph{Math Initial Response.}\nopagebreak[4]
\promptbox{%
\promptslot{Problem}

Please reason step by step, and put your final answer within \texttt{\textbackslash boxed\{\}}.
}

\newpage
\paragraph{Code Initial Response.}\nopagebreak[4]
\promptbox{%
\promptslot{Problem}

Starter code: \emph{optional}

\promptfence{python}\\
\promptslot{Starter Code}\\
\promptfence{}

You will be given a programming problem. Write a correct Python program that solves it. Return only the code inside a single \promptfence{python} code block.
}

\paragraph{OPD Math Refinement.}\nopagebreak[4]
\promptbox{%
Your task is to rewrite your mathematical solution.

Problem:

\promptslot{Problem}

Your Initial Solution:

\promptslot{Initial Response}

Instructions:

\begin{enumerate}\setlength{\itemsep}{0.12em}\setlength{\topsep}{0.15em}
    \item Preserve the overall structure and reasoning path of your original solution
    \item Identify and fix errors in computation or logic
    \item Keep correct intermediate steps and meaningful work
    \item Output ONLY the rewritten solution
\end{enumerate}
Please reason step by step, and put your final answer within \texttt{\textbackslash boxed\{\}}.
}

\paragraph{OPSD Math Refinement.}\nopagebreak[4]
\promptbox{%
Your task is to rewrite your mathematical solution using the reference solution as guidance.

Problem:

\promptslot{Problem}

Reference Solution:

\promptslot{Expert Solution}

Your Initial Solution:

\promptslot{Initial Response}

Instructions:

\begin{enumerate}\setlength{\itemsep}{0.12em}\setlength{\topsep}{0.15em}
    \item Review the reference solution to understand the target reasoning and method
    \item Rewrite your solution so it is consistent with the reference solution
    \item Keep useful parts of your original structure and style when appropriate
    \item Output ONLY the rewritten solution
\end{enumerate}
Please reason step by step, and put your final answer within \texttt{\textbackslash boxed\{\}}.
}

\newpage
\paragraph{OPD Code Refinement.}\nopagebreak[4]
\promptbox{%
Your task is to rewrite your Python solution.

Problem:

\promptslot{Problem}

Your Initial Solution:

\promptslot{Initial Response}

Instructions:

\begin{enumerate}\setlength{\itemsep}{0.12em}\setlength{\topsep}{0.15em}
    \item Fix correctness issues and edge cases
    \item Preserve useful parts of the original approach when appropriate
    \item Output ONLY the rewritten Python solution
\end{enumerate}
Return only the corrected Python code inside a single \promptfence{python} code block.
}

\paragraph{OPSD Code Refinement.}\nopagebreak[4]
\promptbox{%
Your task is to rewrite your Python solution using the reference solution as guidance.

Problem:

\promptslot{Problem}

Reference Solution:

\promptfence{python}\\
\promptslot{Expert Solution}\\
\promptfence{}

Your Initial Solution:

\promptslot{Initial Response}

Instructions:

\begin{enumerate}\setlength{\itemsep}{0.12em}\setlength{\topsep}{0.15em}
    \item Fix correctness issues and edge cases
    \item Preserve useful parts of the original approach when appropriate
    \item Output ONLY the rewritten Python solution
\end{enumerate}
Return only the corrected Python code inside a single \promptfence{python} code block.
}

\end{document}